%% file: main.tex
\documentclass[10pt,twocolumn,letterpaper]{article}

\usepackage[pagenumbers]{lumos} %

\definecolor{lumosblue}{rgb}{0.21,0.49,0.74}
\usepackage[pagebackref,breaklinks,colorlinks,citecolor=lumosblue,bookmarks=false]{hyperref}
\newcommand\blfootnote[1]{%
  \begingroup
  \renewcommand\thefootnote{}\footnote{#1}%
  \addtocounter{footnote}{-1}%
  \endgroup
}
\usepackage{multirow}
\usepackage{makecell} 
\usepackage{animate}
\usepackage{graphicx}
\usepackage{xcolor}
\usepackage[misc]{ifsym}
\usepackage{footnote}

\input{preamble}

\title{Learning Visual Generative Priors without Text}

\author{
Shuailei Ma$^1$$^*$, Kecheng Zheng$^{2}$$^*$, Ying Wei$^1$\textsuperscript{\Letter}, Wei Wu$^2$, Fan Lu$^2$, \\ 
Yifei Zhang$^3$, Chen-Wei Xie$^4$, Biao Gong$^{2}$, Jiapeng Zhu$^5$, Yujun Shen$^{2}$\textsuperscript{\Letter}\\
$^1$ College of Information Science and Engineering, Northeastern University, Shenyang 110819, China \\
$^2$ Ant Group  \quad
$^3$ Shanghai Jiao Tong University \quad
$^4$ Alibaba Group \quad
$^5$ HKUST \\
\tt\small
\href{https://ant-research.github.io/lumos}{https://ant-research.github.io/lumos}
}

\begin{document}
\input{sec/teaser}
\blfootnote{\small *~: Equal contribution. \Letter~: Corresponding author.}

\input{sec/0_abstract}    
\input{sec/1_intro}
\input{sec/2_related_work}

\input{sec/3_method}
\input{sec/4_exp}

\input{sec/5_conclusion}
\input{sec/Acknowledgement}

\input{sec/6_ref} 
\input{sec/7_appendix}

\end{document}

%% file: preamble.tex
\usepackage{graphicx} 
\usepackage{subcaption}
\usepackage{orcidlink}
\usepackage{url}
\usepackage{wrapfig}
\usepackage{color, colortbl}
\usepackage{physics}
\usepackage{multirow}
\usepackage{xcolor}
\usepackage{footnote}
\usepackage{float}
\usepackage{lipsum}
\usepackage[hypcap=false]{caption}
\usepackage{tabularx,booktabs}
\usepackage{etoolbox}
\usepackage{threeparttable}
\usepackage{amsmath, amssymb, caption, multirow, overpic, textpos}
\usepackage{tabularray}
\usepackage{titletoc}
\usepackage{txfonts}
\newcolumntype{x}[1]{>{\centering\arraybackslash}p{#1pt}}
\newcolumntype{y}[1]{>{\raggedright\arraybackslash}p{#1pt}}
\newcolumntype{z}[1]{>{\raggedleft\arraybackslash}p{#1pt}}
\newcommand{\red}[1]{\textcolor{red}{#1}}

\input{def}
\newlength\savewidth

\definecolor{baselinecolor}{gray}{.9}

\newcommand{\tocite}[1]{{\color{red} [TO CITE]}}
\newcommand{\methodname}{Lumos}
\newcommand{\method}{\texttt{\methodname}\xspace}
\newcommand{\supp}{\textit{Supplementary Material}\xspace}

%% file: def.tex
\DeclareMathAlphabet\mathbfcal{OMS}{cmsy}{b}{n}

\def\0{{\bf 0}}
\def\1{{\bf 1}}

\def\eg{\emph{e.g.}} 
\def\ie{\emph{i.e.}}

\usepackage{amsmath}

%% file: sec/teaser.tex
\twocolumn[{
\maketitle
\begin{center}
    \vspace{-20pt}
    \includegraphics[width=0.95\linewidth]{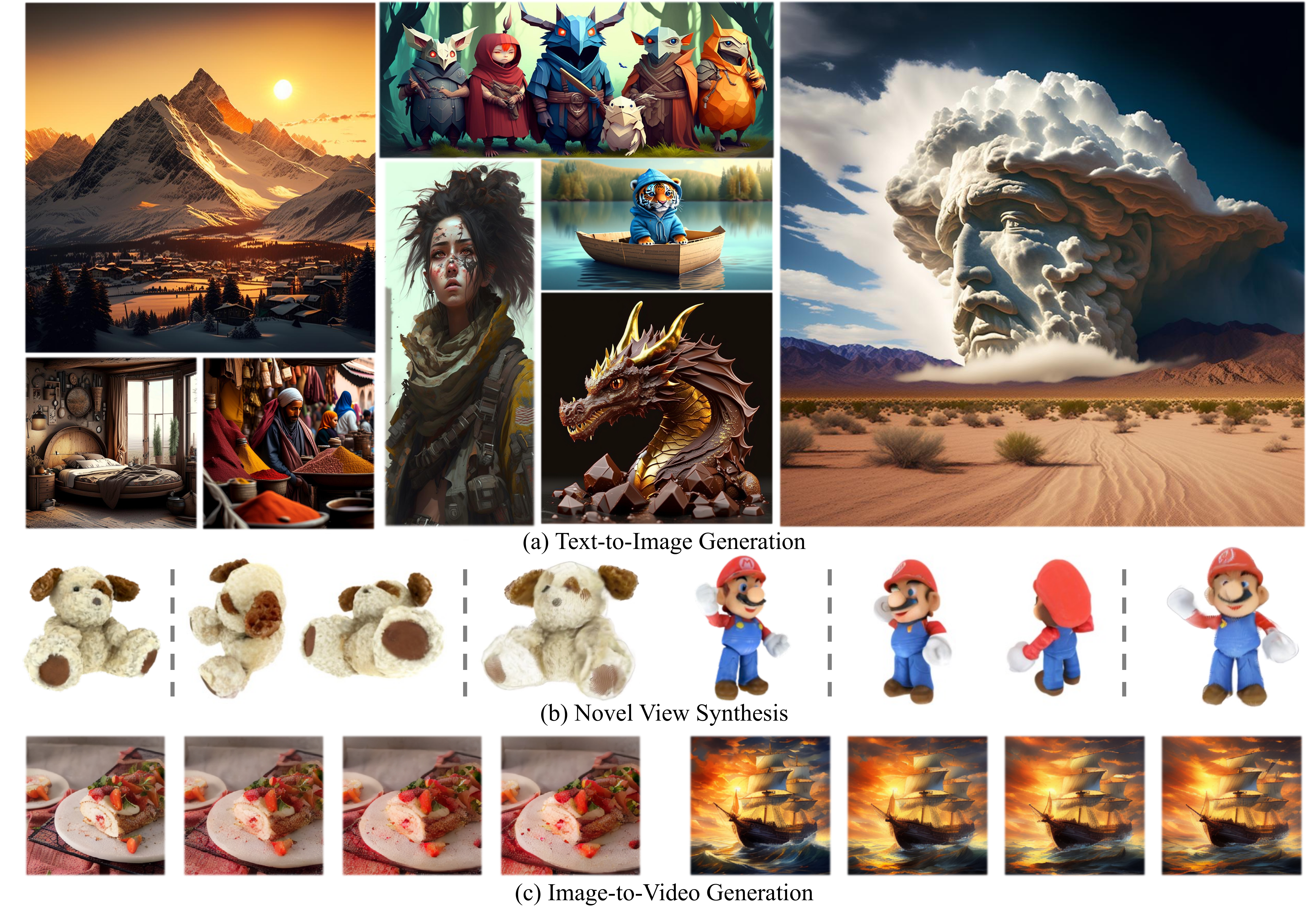} %
    \vspace{-10pt}
    \captionof{figure}{%
        \textbf{Diverse downstream tasks} of \method including (a) text-to-image generation, (b) novel view synthesis (left: input view, middle: random novel views, right: reconstruction Gaussian) and (c) image-to-video generation.
    }
    \label{fig:teaser}
    \vspace{-6pt}
\end{center}
}]

%% file: sec/0_abstract.tex
\begin{abstract}
Although text-to-image (T2I) models have recently thrived as visual generative priors, their reliance on high-quality text-image pairs makes scaling up expensive.
We argue that grasping the cross-modality alignment is \textbf{not} a necessity for a sound visual generative prior, whose focus should be on texture modeling.
Such a philosophy inspires us to study image-to-image (I2I) generation, where models can learn from in-the-wild images in a self-supervised manner.
We first develop a pure vision-based training framework, \method, and confirm the feasibility and the scalability of learning I2I models.
We then find that, as an upstream task of T2I, our I2I model serves as a more foundational visual prior and achieves on-par or better performance than existing T2I models using only 1/10 text-image pairs for fine-tuning.
We further demonstrate the superiority of I2I priors over T2I priors on some text-irrelevant visual generative tasks, like image-to-3D and image-to-video.

\end{abstract}
\vspace{-15pt}

%% file: sec/1_intro.tex
\section{Introduction}
\label{sec:intro}
Text-to-Image (T2I) generative models~\cite{dalle,dalle2,dalle3,sd3,ldm_sd}, which showcase the stunning ability to generate high-fidelity images from a given text prompt, have made a remarkable leap in the evolution of visual generation.
These well-trained T2I generative models are usually regarded as visual generative priors for downstream visual synthesis tasks, such as image-to-video (I2V)~\cite{svd} and novel view synthesis (NVS)~\cite{zero123}, endowing the downstream models with rich semantic information.

\begin{figure}[t]\centering
\subfloat[Noise data for T2I.\label{fig:intro_a}]
{\includegraphics[width=0.45\linewidth]{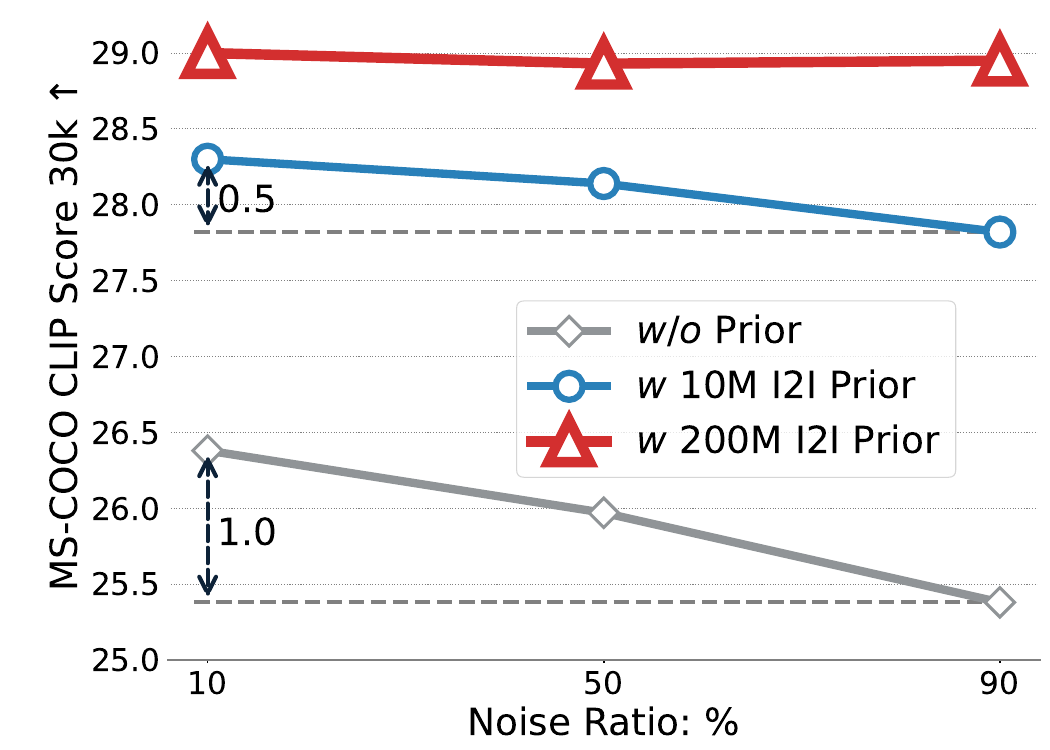}}\hfill
\subfloat[Training curve of scaling.\label{fig:intro_b}]
{\includegraphics[width=0.45\linewidth]{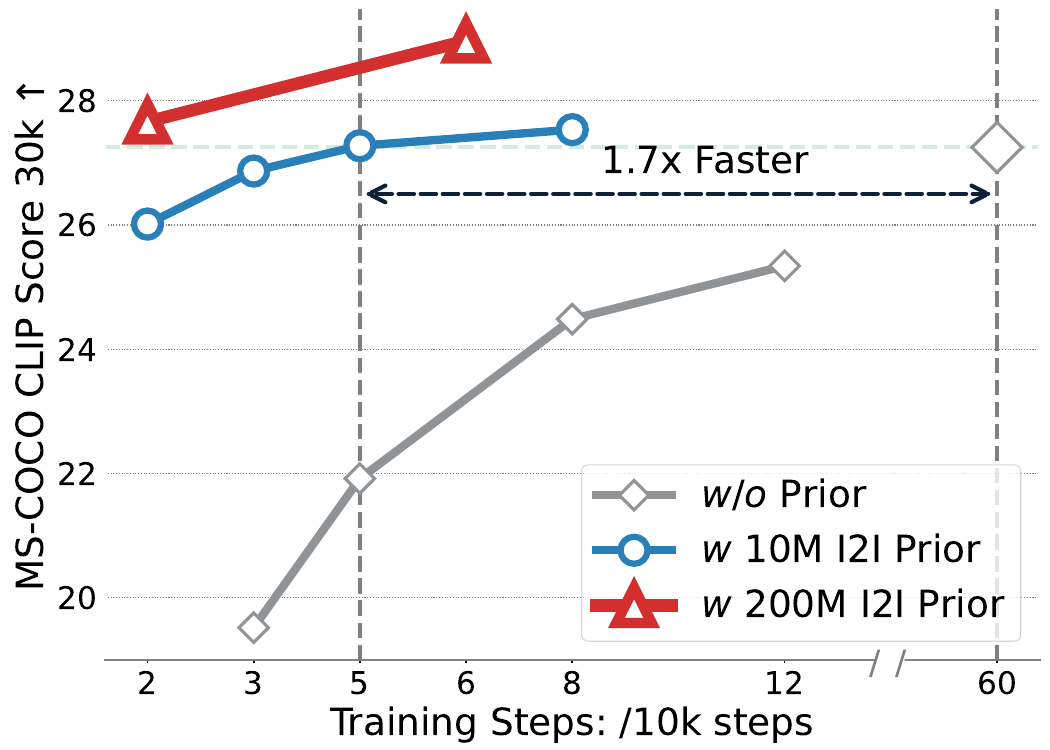}}\\
\subfloat[Different I2I priors.\label{fig:intro_c}]
{\includegraphics[width=0.45\linewidth]{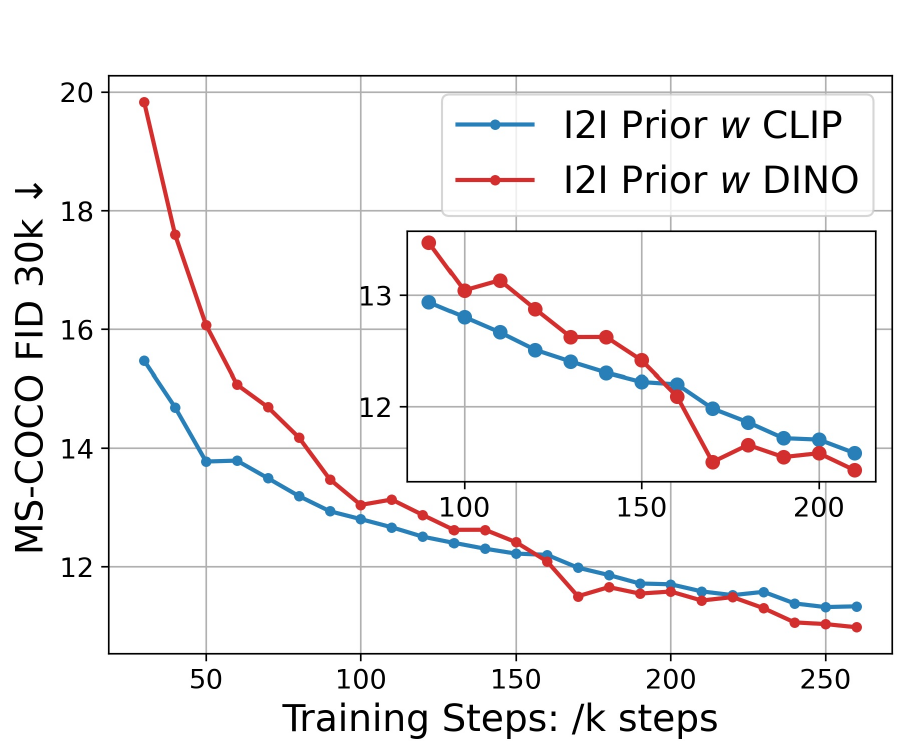}}\hfill
\subfloat[Text-irrelevant visual tasks.\label{fig:intro_d}]
{\includegraphics[width=0.45\linewidth]{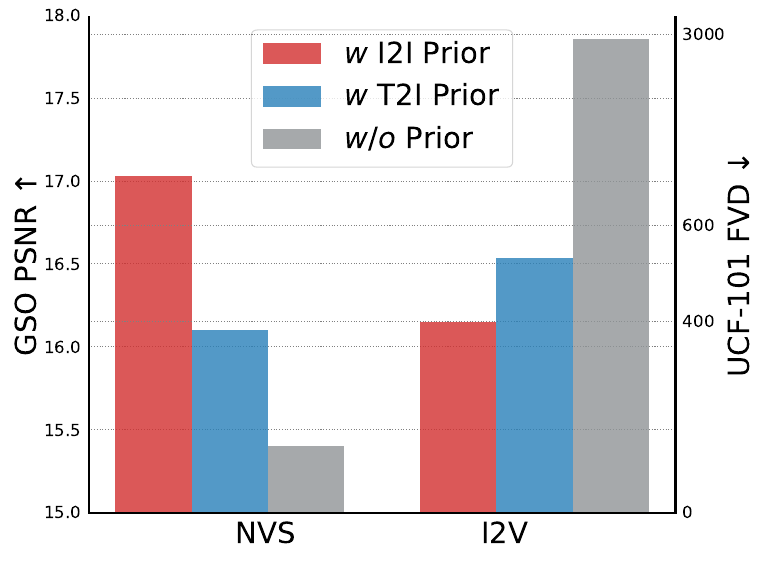}}
\vspace{-8pt}
\caption{
    \textbf{Various image generation tasks can be improved by our image-to-image priors.} 
    I2I prior enables the downstream T2I model to decrease dependence on high-quality data, and with data scaling up in I2I, it enjoys a larger performance improvement.
    We also adopt a pure vision-based I2I generation (\ie, I2I prior with DINO) that is a late bloomer for T2I generation.
    We further demonstrate the superiority of I2I priors over T2I priors on some text-irrelevant vision tasks, like I2V and NVS.
}
\vspace{-8pt}
\label{fig:intro}
\end{figure}

However, recent research~\cite{pixart-a,qihoo-t2x} has demonstrated that a well-trained T2I generative model intensely depends on high-quality image-text pairs.
As shown in~\Cref{fig:intro_a}, with the noise ratio of text-image pairs increasing from 10\% to 90\%, the performance of the T2I model degrades around 1.0 CLIP score.
It indicates that the performance of T2I model is sensitive to the quality of text-image pairs, because T2I models simultaneously focus on two difficult issues: learning \textit{texture modeling} and \textit{text-image alignment}.
Noisy texts of images may be harmful to \textit{text-image alignment} and further exacerbate the difficulty of \textit{texture modeling} in generative models.
Scaling up the well-aligned image-text paired data can address this issue but is quite expensive.
As a result, it is a necessary step towards unleashing the power of large-scale unannotated data.
A commonly adopted strategy is to utilize unconditional generative methods~\cite{dalle2,RCG2023} to model texture, which can learn the probabilistic distributions of in-the-wild data without requiring paired data.
With this observation, we argue that grasping the cross-modality alignment is \textbf{not} a necessity for a sound visual generative prior, whose focus should be on texture modeling.

Such a philosophy inspires us to study image-to-image (I2I) generation, where models can learn from in-the-wild images in a self-supervised manner.
We first develop a pure vision-based framework, termed as \method, to generate images conditioned on image features extracted by a pre-trained vision encoder (\eg, DINO).
Then, we fine-tune \method to various downstream visual synthesis tasks (\eg, T2I).
As shown in~\Cref{fig:intro_a}, thanks to I2I prior, our T2I model shows less reliance on high-quality data, exhibiting 0.5 less degradation in CLIP score compared with \textit{w/o} I2I prior.
Moreover, as illustrated in~\Cref{fig:intro_b}, with the data of I2I scaling up from 10M to 200M, our model achieves better performance, confirming the feasibility and the scalability of learning I2I priors.
We hypothesize that the performance gap arises because the I2I priors enhance texture modeling, thereby simplifying the T2I generative process.
Another interesting question lies in the influence of image encoders on I2I model.
For a T2I model, text-image alignment is crucial. Therefore, a straightforward approach would be to use a text-image-aligned CLIP in the I2I model, which may enhance the T2I model's performance compared to DINO.
However, as shown in~\Cref{fig:intro_c}, although CLIP shows greater advantages in the early stages of learning I2I priors, DINO, as a late bloomer, achieves better COCO-30K FID than CLIP in the final steps.
It demonstrates the feasibility of the pure vision-based training framework.
Based on these, we find that, as an upstream task of T2I, our pure-vision I2I model serves as a more foundational visual prior.

We further demonstrate the superiority of I2I priors over T2I priors on some text-irrelevant vision tasks, like image-to-3D and image-to-video.
As shown in~\Cref{fig:intro_d}, we observe that I2I priors bring better performance than T2I priors. 
On these text-irrelevant vision tasks, it is hard to design a suitable text prompt.
However, for I2I prior, it naturally does not require text input and can be efficiently transferred directly to these tasks.
Thus, our I2I prior can achieve better performance than T2I prior, which can serve as a more foundational visual prior. 
We hope our method, results, and analysis will encourage future research on pure visual generative priors.

%% file: sec/2_related_work.tex
\section{Related Work}\label{sec:related}

\subsection{Unsupervised Representation Learning}
Unsupervised pre-training methods~\cite{bert,gpt,simclr,moco,mocov2,dino} are undeniably one of the key contributors to the success of deep learning nowadays, as they break the shackles of expensively supervised data on the deep learning model. 
For Natural Language Processing, masked language modeling and its autoregressive counterparts, \eg, BERT~\cite{bert} and GPT~\cite{gpt}, first demonstrated the potential of unsupervised pre-training. 
Similarly, exploring unsupervised pre-training for visual understanding tasks has never stopped.
Self-supervised learning methods, \eg, MoCo~\cite{moco,mocov2}, simCLR~\cite{simclr}, and DINO~\cite{dino}, are pre-trained in the form of self-distillation without labels, proving that they can learn to extract richer features from images than the supervised counterparts. 
Mask image modeling, \eg, MAE~\cite{he2022masked} and SimMIM~\cite{xie2022simmim} ingeniously adapt the concept of masked modeling from the NLP field to image processing, leading to remarkably improved results in understanding pre-training.
However, the exploration on unsupervised learning in visual generation remains limited.
In this work, we investigate developing an unsupervised visual generative learning framework, that provides foundational visual priors for various visual synthesis tasks (\eg, text-to-image).

\subsection{Generative Diffusion Models}
Visual generative models have undergone long-term development, targeting learning the probabilistic distributions of data from the images.
Some fundamental methods, such as GAN~\cite{goodfellow2014generative}, VAE~\cite{vae}, and Flow~\cite{flow}, show impressive performance in modeling simple image distributions such as animal face and indoor scene.
In the complex data, generative diffusion models~\cite{dalle3,sd3,ldm_sd} have achieved superior results (\eg, text-to-image generation), whose contents are high-quality.
However, current research~\cite {dalle3,pixart-a} shows that the success of generative diffusion models is inseparable from high-quality text-image pairs, which greatly limits their scalability. 
To address this problem, PixArt-$\alpha$~\cite{pixart-a} mentions that class-to-image pre-training on ImageNet~\cite{deng2009imagenet} can serve as a pre-trained model for text-to-image generation, accelerating the convergence speed of the model. 
But still, human annotations are expensive and hard to scale up.
Inspired by unconditional generation, DALLE-2~\cite{dalle2} and RCG~\cite{RCG2023} proposed to use the image-to-image model as an intermediate bridge between the downstream models (\ie, T2I) and the unconditional generation diffusion model. 
In this paper, we explore a novel I2I generative method that captures pure visual priors from in-the-wild images, and confirm its feasibility and scalability.

%% file: sec/3_method.tex
\section{Method}\label{sec:method}
\begin{figure*}[htbp]
    \centering
    \includegraphics[width=0.9\textwidth]{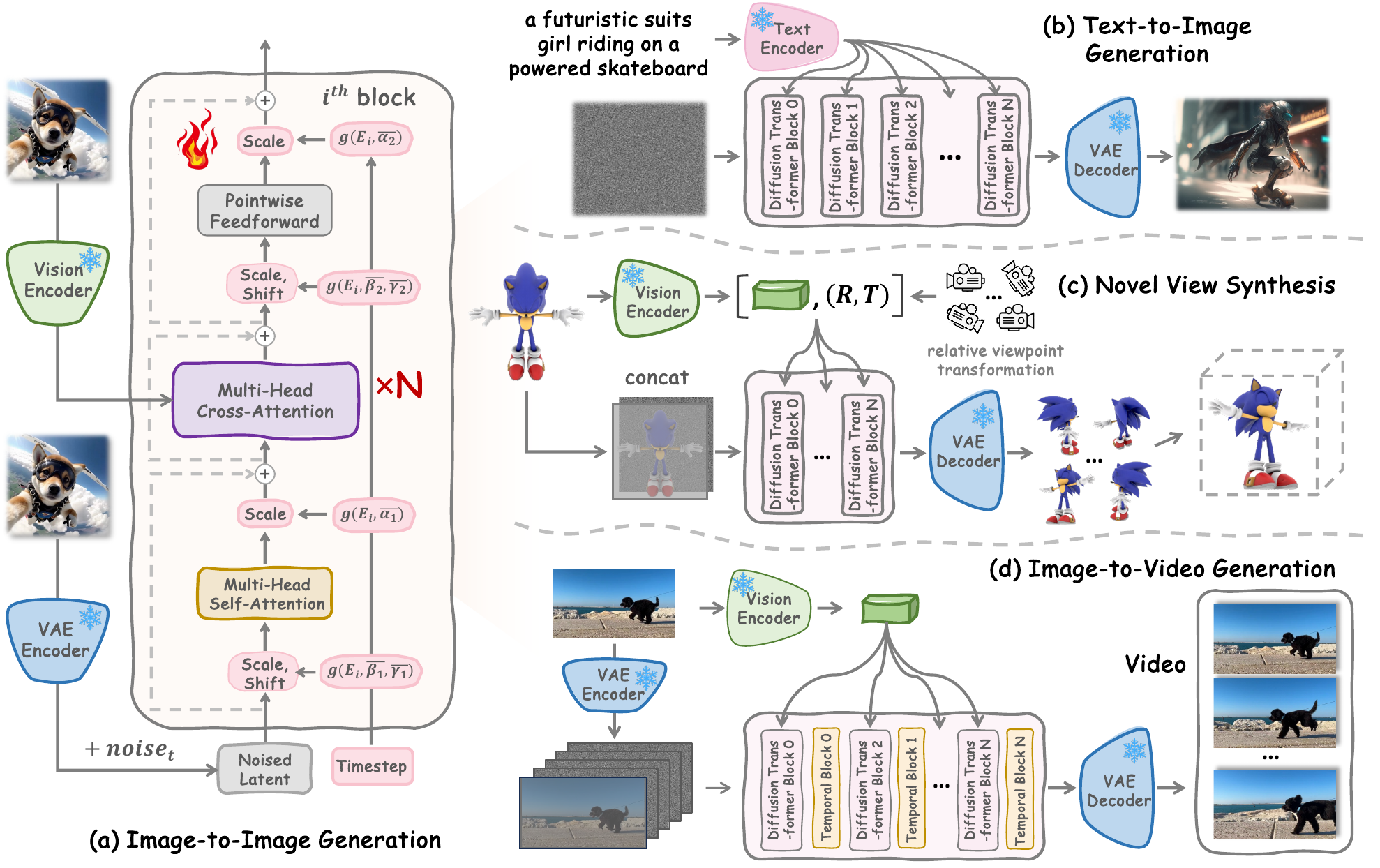}
    \vspace{-10pt}
    \caption{\textbf{Overall architecture of our framework.} (a) Image-to-Image Generation, (b) Text-to-Image Generation, (c) Novel View Synthesis and (d) Image-to-Video Generation.}
    \vspace{-10pt}
    \label{method}
\end{figure*}
In this section, we firstly introduce the image-to-image generation to learn from in the wild images in the self-supervised manner. Then, we transfer the well-trained I2I model, as the image-to-image prior, to some downstream vision tasks. 

\subsection{Image-to-Image Generative Model}
Given an image $x \in \mathbb{R}^{h \times w \times 3}$, we leverage the pre-trained autoencoder~\cite{ldm_sd} $\mathcal{E}$ to get its latent representation $z = \mathcal{E}(x)$.
Then, we utilize the off-the-shelf vision encoder (\eg, DINO)~\cite{clip,dino,moco,mocov2} $\tau^{\text{image}}$ to extract its visual semantic feature $\tau^{\text{img}}(x) \in \mathbb{R}^{M \times d}$.
In this way, the pre-trained vision encoders, achieving state-of-the-art performance in representation learning, can provide rich semantic content, which is crucial for guiding image generation.
Meanwhile, it is simple enough to scale up.
As shown in \cref{method} (a), condition on the resulting semantic features, we then learn the image-to-image latent diffusion model $\epsilon_{\theta_{\text{I2I}}}$ via
\begin{equation}\label{eq1}
\mathcal{L}_{\theta_{\text{I2I}}}=\mathbb{E}_{\mathcal{E}(x), x, \epsilon, t}\left[\left\|\epsilon-\epsilon_\theta\left(z_t, t, \tau^{\text{img}}(x)\right)\right\|_2^2\right],
\end{equation}
where $\epsilon \sim \mathcal{N}(\boldsymbol{0},\boldsymbol{I})$ and both of $\mathcal{E}$ and $\tau^{\text{img}}$ remain frozen throughout the pre-training process.

We use DiT~\cite{dit} as the vision backbone.
We conduct ablation experiments to explore different types of vision encoders (\eg, pure-vision pre-training $vs.$ language-image pre-training), and {visual semantic condition types (global feature $vs.$ local features)}. We briefly summarize our observations as following:
\begin{itemize}
    \item \textit{Pure-vision visual encoder is better than the language image pre-training encoder for I2I generation}. Compared to CLIP~\cite{clip}, self-supervised pure-vision visual encoders (\eg, DINO~\cite{dino} and MoCo~\cite{moco,mocov2}) exhibit a faster convergence speed and better performance in terms of the FID metric for image-to-image generation.
    \item \textit{Scaling up the number of images can improve the image generation for I2I generation}. As the data of I2I scales up from 10M to 200M, I2I generative model achieves better performance in FID.
    \item \textit{Using local features as condition decreases the difficulty of I2I generation}. Local features from pre-trained vision encoders, including more detailed content from images, show a significant advantage for I2I generation than global feature.
\end{itemize}
The ablation study is presented in~\cref{ablation study}. 
Based on these observations, I2I generation can utilize the vast abundance of unannotated data to model rich texture.

\subsection{Generative Transfer Learning}
After training a I2I generation model, a fine-tuning stage transfers it to various downstream visual synthesis tasks.
In this section, we regard widely-used text-to-image generation, novel view synthesis, and image-to-video tasks as our downstream tasks.

\noindent\textbf{Text-to-Image Generation.}
Following LDM~\cite{ldm_sd} and PixArt-$\alpha$~\cite{pixart-a}, we use the cross-attention mechanism~\cite{transformer} to attach the condition information from text prompts $y$ (see details in \cref{method} (a) and (b)).
After loading the I2I pre-trained weight $\theta_{\text{I2I}}$, we tune the diffusion model $\epsilon_{\theta_{\text{T2I}}}$ via
\begin{equation}
\mathcal{L}_{\theta_{\text{T2I}}}=\mathbb{E}_{\mathcal{E}(x), y, \epsilon, t}\left[\left\|\epsilon-\epsilon_\theta\left(z_t, t, \tau^{\text{txt}}(y)\right)\right\|_2^2\right],
\end{equation}
where $\tau^{\text{txt}}$ represents the pre-trained text encoder. 
Both $\tau^{\text{txt}}$ and $\mathcal{E}$ are kept frozen during training. 

T2I generation is a challenge task that allows model simultaneously focus on two difficult issues: learning \textit{texture modeling} and \textit{text-image alignment}. Our I2I model can provide the visual generative priors to help texture modeling, and let T2I model pay more attention to text-image alignment. We briefly summarize our observations as following:
\begin{itemize}
    \item \textit{The performance of upstream and downstream is inconsistency}. The performance of downstream tasks is not significantly influenced by the performance (\eg, FID) of I2I model. For example, when using local features as the condition of I2I model, the FID of downstream T2I model is worse than one of global feature.
    \item \textit{Scaling up I2I generation can improve the performance of downstream tasks.} With the data of I2I scaling up from 10M to 200M, our T2I model achieves better performance in terms of FID, confirming the feasibility and the scalability of learning I2I priors.
    \item \textit{Pure-vision visual encoder is better than language-image pre-training models for T2I generation.} Although CLIP shows greater advantages in the early stages of learning I2I priors, DINO, as a late bloomer, achieves better FID than CLIP in the final steps.
\end{itemize}
Based on these observation, we find that, as an upstream task of T2I, our pure-vision I2I model serves as a more foundational visual prior. For the detailed comparison, see the ablation experiments in~\cref{ablation study}.

\noindent\textbf{Text-irrelevant Visual Generative Tasks.}
We further demonstrate the superiority of I2I priors over T2I priors on some text-irrelevant vision tasks, like image-to-3D and image-to-video.
Specifically, following Zero-1-to-3~\cite{zero123}, we tune our I2I model to the novel view synthesis task.
As shown in \cref{method} (c), the task is to synthesize an image of an object from a new camera viewpoint.
Moreover, We also finetune our pre-trained I2I model for the image-to-video generation task, where the video model receives a still input image as the condition as shown in \cref{method} (d).
On these text-irrelevant vision tasks, previous methods adopt pre-trained T2I model as visual generative prior, where it is hard to design a suitable text prompt.
However, for I2I prior, it naturally does not require text input and can be efficiently transferred directly to these tasks.
Our I2I prior can achieve better performance than T2I prior, which can serve as a more foundational visual prior. 

%% file: sec/4_exp.tex
\section{Experiments}\label{sec:exp}

\subsection{Image-to-Image Generation} 
\noindent\textbf{Implementation Details \& Datasets.} 
Since our image-to-image generation does not require labels, we can easily filter totaling 190 million images via some criteria of image quality from existing open-source datasets (\ie, LAION-5B~\cite{laion}, COYO-700M~\cite{coyo-700m}, SAM~\cite{sam}, JourneyDB~\cite{journeydb}, and ImageNet-1K~\cite{deng2009imagenet}). 
More details can refer to \supp.
We use DINO-B as the vision encoder of the image-to-image model and DiT-XL-2~\cite{dit} as the diffusion backbone.
We evaluate the image-to-image model via Fréchet Inception Distance (FID)~\cite{fid} on MSCOCO~\cite{mscoco} and ImageNet-1K~\cite{deng2009imagenet}.

\begin{wraptable}{4}{0.46\linewidth}
    \vspace{-12.4pt}
    \centering
    \scriptsize
    \caption{
        \textbf{Comparison with RCG in I2I generation.}
    }
    \vspace{-3pt}
    \SetTblrInner{rowsep=1.0pt}       %
    \SetTblrInner{colsep=3.0pt}       %
    \label{table:SOTA i2i}
    \begin{tblr}{
        cells={halign=c,valign=m},  %
        cell{1}{1}={r=2}{},         %
        hline{1,3,5}={1.0pt},        %
    }
        Method             & MS-COCO              & ImageNet \\
                           & FID 30K $\downarrow$ & FID 50K $\downarrow$ \\
        RCG~\cite{RCG2023} & 12.70                & 4.89 \\
        \method-I2I               & \textbf{4.82}        & \textbf{2.60} \\
    \end{tblr}
    \vspace{-10pt}
\end{wraptable}
\noindent\textbf{Comparison with SOTA I2I model.}
We compare our model with the open-source state-of-the-art I2I model, \ie, RCG~\cite{RCG2023}, which is pre-trained on ImageNet-1K~\cite{deng2009imagenet}.
For the RCG model, we use the open-source version, namely pre-trained DiT-XL/2 (the same as our diffusion model structure) conditioned on Moco-v3 ViT-B~\cite{moco}. 
As shown in~\Cref{table:SOTA i2i}, \method significantly outperforms RCG in terms of FID on ImageNet-1K and MSCOCO, demonstrating the excellent scalability of Image-to-Image generation on large-scale datasets.

\subsection{Downstream Visual Generative Tasks} 
Subsequently, \method-I2I is used as visual generative prior and transferred to various downstream tasks, including Text-to-Image Generation, Novel View Synthesis, and Image-to-Video Generation.

\subsubsection{Text-to-Image Generation}
We first transfer \method-I2I to the text-to-image generation task and compare \method-T2I with the state-of-the-art T2I models in terms of the quality of the generated images and alignment assessment.

\begin{table}[t]
    \centering
    \small
    \caption{\textbf{Comparison to the recent text-to-image models on COCO FID-30k.} ` T\&I Pairs ' refers to the number of the training text-image pairs. $^{\star}$ refers to the result of ` PixArt-$\alpha$-256 '. $^{\dagger}$ indicates evaluating with the long captions. Both + and - in the table denote the unknown internal dataset size and training step.}
    \vspace{-2mm}
    \SetTblrInner{rowsep=0.4pt}      
    \SetTblrInner{colsep=6.3pt}      %
    \begin{tblr}{
        cells={halign=c,valign=m},   %
        column{1}={halign=l},
        hline{1,2,16,18}={1.0pt},         %
        vline{2}={1-17}{},
    }
        Method                                      & Type & T\&I Pairs & Steps & FID-30K$\downarrow$ \\ 
        DALL$\cdot$E~\cite{dalle}                   & Diff & 250M       & -     & 27.50   \\
        GLIDE~\cite{glide}                          & Diff & 250M       & 2500k & 12.24   \\
        LDM~\cite{ldm_sd}                           & Diff & 400M       & -     & 12.64   \\
        DALL$\cdot$E2~\cite{dalle2}                 & Diff & 650M       & 2400k & 10.39   \\
        SDv1.5~\cite{ldm_sd}                        & Diff & 2000M      & 1026k & 9.62    \\
        GigaGAN~\cite{gigagan}                      & GAN  & 2700M      & 1350k & 9.09    \\
        Imagen~\cite{imagen}                        & Diff & 860M       & 5000k & 7.27    \\
        RAPHAEL~\cite{raphael}                      & Diff & 5000M+     & -     & 6.61    \\
        PixArt-$\alpha$~\cite{pixart-a}             & Diff & 24M        & 240k  & 7.32   \\
        PixArt-$\alpha^{\star}$~\cite{pixart-a}     & Diff & 24M        & 240k  & 23.67   \\
        PixArt-$\alpha^{\star\dagger}$~\cite{pixart-a}   & Diff & 24M        & 240k  & 21.35   \\
        F1ux.1~\cite{flux}                          & Diff & -          & -     & 22.76   \\
        Kolors~\cite{kolors}                        & Diff & -          & -     & 23.15   \\
        Qihoo-T2I~\cite{qihoo-t2x}                  & Diff & 50M        & 100k  & 15.70   \\
        \method-T2I                                 & Diff & 30M        & 65k   & 12.20  \\
        \method-T2I$^{\dagger}$                     & Diff & 30M        & 65k   & \textbf{6.44}    \\ 
    \end{tblr}
    \vspace{-10pt}
    \label{tab:SOTA_t2i}
\end{table}

\noindent\textbf{Implementation Details \& Datasets.} 
We construct a dataset of 30 million text-image pairs, all of which are sourced from easily accessible open datasets. This includes 10M and 5M images selected from LAION-5B and COYO-700M, respectively. Additionally, 10M from SAM~\cite{sam}, 4M from JourneyDB~\cite{journeydb}, and 1M from Imagenet-1K~\cite{deng2009imagenet} are selected. Following DALLE-3~\cite{dalle3}, we incorporate raw captions and detailed captions generated by InternVL~\cite{internvl} into training. Besides FID, we evaluate T2I models on GenEval~\cite{geneval} and DPG-Bench~\cite{dpgbench}. We use the T5~\cite{T5} (specifically 4.3B Flan-T5-XXL) as the text encoder of our T2I model. More details are in~\supp.

\noindent\textbf{Quantitative Comparison for FID score.} 
In~\Cref{tab:SOTA_t2i}, we report the evaluation results on the widely used MS-COCO dataset in terms of FID score and compare \method with existing SOTA T2I models.
It is worth mentioning that we evaluate the open-sourced model of PixArt-$\alpha$~\cite{pixart-a} that is trained on all the datasets proposed in their paper.
Thanks to the visual generative prior from the pre-trained image-to-image model, \method-T2I model achieves on-par or better performance than existing T2I models, requiring only a minimal amount of training data and steps.
Moreover, \method-T2I surpasses all current text-to-image models in terms of the FID metric with the help of long captions.
The experimental results demonstrate the significant benefits of the I2I prior for T2I models.
\par

\begin{table*}[htbp]
    \centering
    \caption{
        \textbf{Text-image alignment comparison with the state-of-the-art text-to-image models on GenEval~\cite{geneval} and DPG-Bench~\cite{dpgbench}.} 
        Our model is decently trained on publicly available datasets without using any self-collected data.
        We highlight the \textbf{best} and \textit{second best}.
    }
    \vspace{-2mm}
    \centering
    \small
    \SetTblrInner{rowsep=0.5pt}      
    \SetTblrInner{colsep=1pt}      %
    \resizebox{\textwidth}{!}{
    \begin{tblr}{
        cells={halign=c,valign=m},   %
        column{1}={halign=l},
        cell{1}{1,2}={r=3}{},
        cell{1}{3}={c=7}{},
        cell{1}{10}={c=6}{},
        cell{2}{3}={c=2}{},
        cell{2}{5,6,8,9,10,11,12,13,14,15}={r=2}{},
        hline{1,4,10,15,16}={1.0pt},         %
        hline{2,3}={1-20}{},
        vline{9,15}={2-20}{},
        vline{2,3,10}={1-20}{},
    }
        Model & Param & GenEval~\cite{geneval}  & &  &        &              &          &         & DPG-Bench~\cite{dpgbench} \\
              &           & Object &     & Counting & Colors &     Color    & Position & \textbf{Overall}$\uparrow$ & Global & Entity & Attribute & Relation & Other & \textbf{Average}$\uparrow$ \\
              &           & Single & Two &          &        &  Attribution &          &         &        &        &           &          &       &         \\
        {\color{gray} LUMINA-Next~\cite{zhuo2024lumina}} & {\color{gray} 2.0B} & {\color{gray} 0.92} & {\color{gray} 0.46} & {\color{gray} 0.48} & {\color{gray} 0.70} & {\color{gray} 0.13} & {\color{gray} 0.09} & {\color{gray} 0.46} & {\color{gray} 82.8} & {\color{gray} 88.7} & {\color{gray} 86.4} & {\color{gray} 80.5} & {\color{gray} 81.8} & {\color{gray} 74.6} \\
        
        {\color{gray} SDXL~\cite{sdxl}} & {\color{gray} 2.6B} & {\color{gray} 0.98} & {\color{gray} 0.74} & {\color{gray} 0.39} & {\color{gray} 0.85} & {\color{gray} 0.23} & {\color{gray} 0.15} & {\color{gray} 0.55} & {\color{gray} 83.3} & {\color{gray} 82.4} &	{\color{gray} 80.9}	& {\color{gray} 86.8} &	{\color{gray} 80.4} & {\color{gray} 74.7}\\

        {\color{gray} Playground v2.5~\cite{li2024playground}} & {\color{gray} 2.6B} & {\color{gray} 0.98} & {\color{gray} 0.77} & {\color{gray} 0.52} & {\color{gray} 0.84} & {\color{gray} 0.17} & {\color{gray} 0.11} & {\color{gray} 0.56} & {\color{gray} 83.1} & {\color{gray} 82.6} & {\color{gray} 81.2} & {\color{gray} 84.1} & {\color{gray} 83.5} & {\color{gray} 75.5} \\
        
        {\color{gray} SD3-8B~\cite{sd3}} & {\color{gray} 8.0B} & {\color{gray} 0.99} & {\color{gray} 0.94} & {\color{gray} 0.72} & {\color{gray} 0.89} & {\color{gray} 0.60} & {\color{gray} 0.33} & {\color{gray} 0.74} & {\color{gray} -} & {\color{gray} -} & {\color{gray} -} & {\color{gray} -} & {\color{gray} -} & {\color{gray} -} \\
        
        {\color{gray} FLUX-dev~\cite{flux}} & {\color{gray} 12.0B} & {\color{gray} 0.99} & {\color{gray} 0.81} & {\color{gray} 0.79} & {\color{gray} 0.74} & {\color{gray} 0.47} & {\color{gray} 0.20} & {\color{gray} 0.67} & {\color{gray} 82.1} & {\color{gray} 89.5} &	{\color{gray} 88.7} & {\color{gray} 91.1} & {\color{gray} 89.4}	& {\color{gray} 84.0}\\
        {\color{gray} FLUX-schnell~\cite{flux}} & {\color{gray} 12.0B} & {\color{gray} 0.99} & {\color{gray} 0.92} & {\color{gray} 0.73} & {\color{gray} 0.78} & {\color{gray} 0.54} & {\color{gray} 0.28} & {\color{gray} 0.71} & {\color{gray} 91.2}	& {\color{gray} 91.3} & {\color{gray} 89.7}& {\color{gray} 86.5} & {\color{gray} 87.0} & {\color{gray} 84.8} \\
        SDv1.5~\cite{ldm_sd} & 0.9B & 0.97 & 0.38 & 0.35 & 0.76 & 0.06 & 0.04 & 0.43 & 74.6 & 74.2 & 75.4 & 73.5 & 67.8 & 63.2\\
        SDv2.1~\cite{ldm_sd} & 0.9B & \underline{0.98} & 0.51 & 0.44 & \textbf{0.85} & 0.17 & 0.07 & 0.50 & 77.7 & 78.1 & 74.9 & 80.7 & 80.7 & 68.1\\
        PixArt-$\alpha$~\cite{pixart-a} & 0.6B & \underline{0.98} & 0.50 & 0.44 & 0.80 & 0.07 & 0.08 & 0.48 & 81.7 & 80.1 &	\underline{80.4} & 81.7 & 76.5 & 71.6 \\
        PixArt-$\Sigma$~\cite{pixart-s} & 0.6B & \underline{0.98} & 0.59 & \underline{0.50} & 0.80 & 0.15 & 0.10 & 0.52 & \textbf{87.5} & \underline{87.1} &	\textbf{86.5} & \underline{84.0} & \underline{86.1} & \underline{79.5} \\
        SD3-1B~\cite{sd3} & 1.0B & 0.97 & \textbf{0.72} & \textbf{0.52} & 0.78 & \textbf{0.34} & \textbf{0.16} & \textbf{0.58} & - & - & - & - & - & - \\ 
        \method-T2I & 0.8B & \textbf{0.99} & \underline{0.64}  & \textbf{0.52} & \underline{0.84} & \underline{0.30} & \underline{0.15} & \underline{0.57} & \underline{87.4} & \textbf{87.4} & \textbf{86.5} & \textbf{87.2} & \textbf{88.1} & \textbf{79.9} \\ 
    \end{tblr}}
    \vspace{-10pt}
    \label{tab:SOTA_Geneval}
\end{table*}

\begin{figure}[htbp]
\centering
    \subfloat[LPIPS]{
     \centering
     \includegraphics[width=0.31\linewidth]{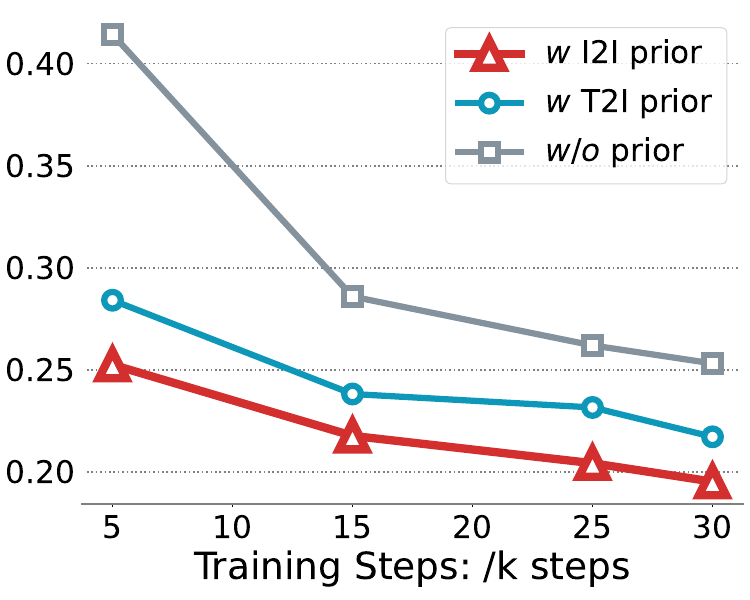}
     }
    \subfloat[PSNR]{
         \centering
         \includegraphics[width=0.31\linewidth]{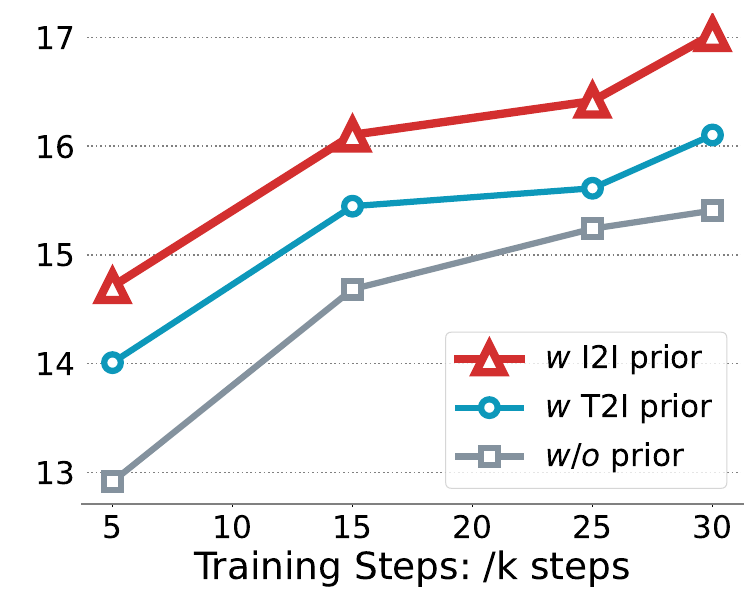}
    }
    \subfloat[SSIM]{
         \centering
         \includegraphics[width=0.31\linewidth]{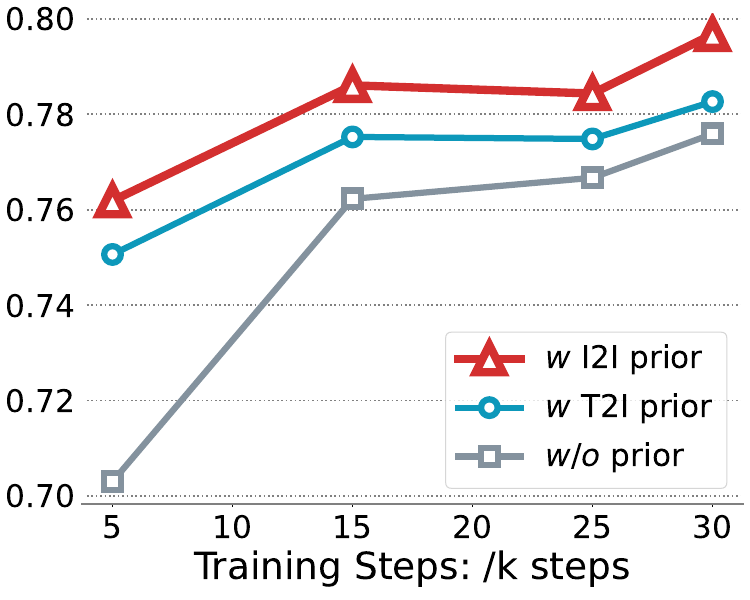}
    }
    \vspace{-2mm}
    \captionof{figure}{\textbf{Comparison with different priors for novel view synthesis.} I2I prior shows better metrics consistently from the start of fine-tuning.}
    \label{fig:ablate_nvs}
    \vspace{-2mm}
\end{figure}

\noindent\textbf{Quantitative Comparison for Image-text Alignment Assessment.} 
Beyond the image fidelity evaluation, we evaluate the alignment between the generated images and text prompts through GenEval~\cite{geneval} and DPG-Bench~\cite{dpgbench}. 
GenEval~\cite{geneval} is designed for evaluating T2I models from multiple perspectives, including the number of objects, color, attributes, and position. 
DPG-Bench~\cite{dpgbench} is a more challenging benchmark, which evaluates the generating performance of the dense prompts and the relation between the entities. 
Despite using only a small amount of training data and steps, the ability of \method-T2I's text-image alignment is on par with or exceeds existing models with a similar parameter scale, as shown in the comparison results of Table.~\ref{tab:SOTA_Geneval}. And on the more challenging DPG-Bench~\cite{dpgbench}, \method-T2I's performance is still outstanding.
This further proves the feasibility of I2I pre-training followed by T2I alignment training framework in the text-to-image task.
\par

\begin{figure*}[htbp]
    \centering
    \includegraphics[width=0.95\textwidth]{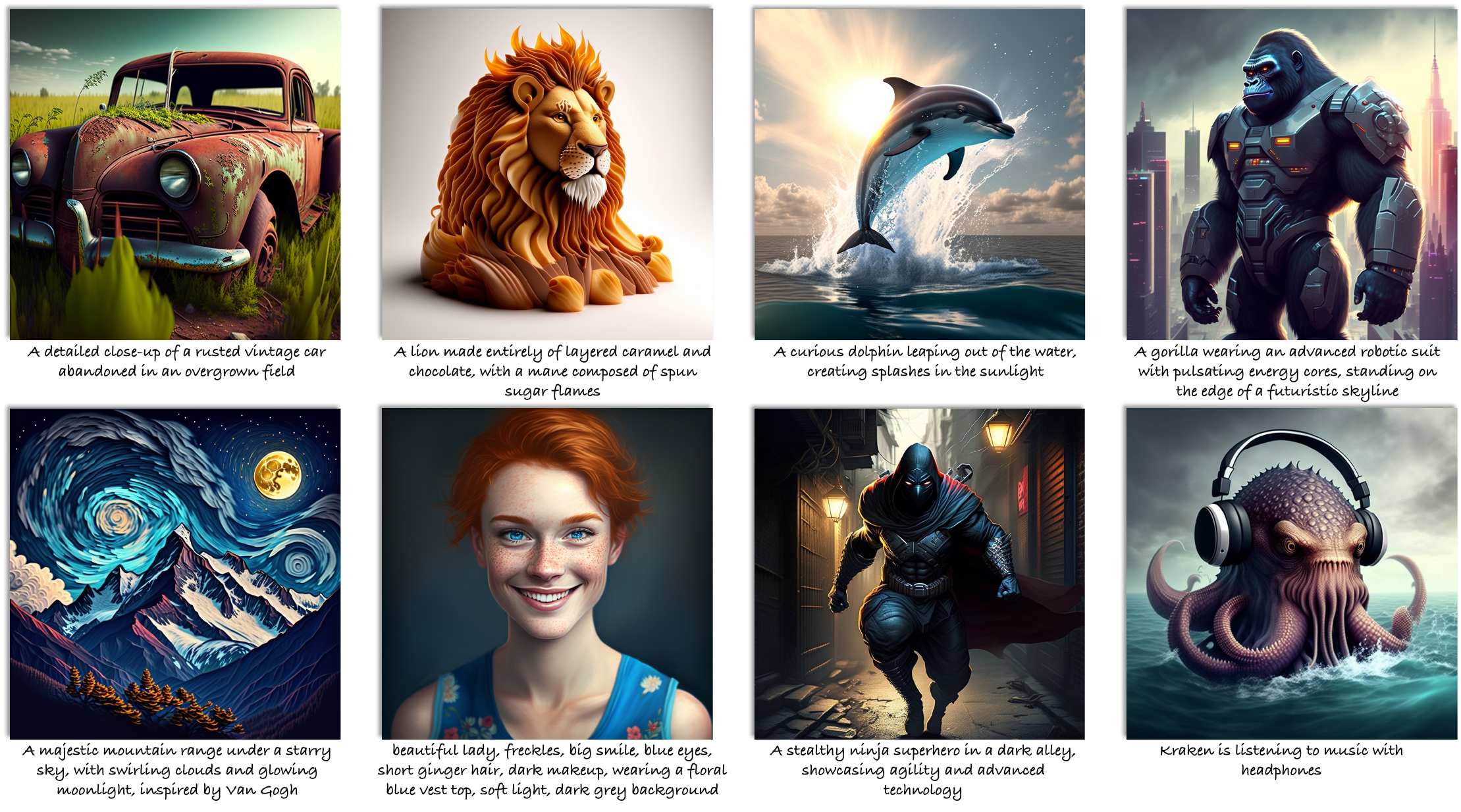}
    \vspace{-10pt}
    \caption{Samples produced by \method-T2I exhibit exceptional quality, characterized by a remarkable level of fidelity and precision in adhering to the provided textual prompts.}
    \vspace{-12pt}
    \label{fig:qualitative}
\end{figure*}

\noindent\textbf{Qualitative Results.} We provide more qualitative generated images as shown in
~\Cref{fig:qualitative} 
to validate the alignment between the generated images and prompts.

\subsubsection{Novel View Synthesis}
We further validate the advantages of the I2I prior on 3D tasks through the novel view synthesis task.\par
\noindent\textbf{Implementation Details \& Datasets.}
We transfer \method-I2I to novel view synthesis through a subset (750k) of the released Objaverse~\cite{objaverse} dataset. Following Zero123~\cite{zero123}, we randomly sample 32 camera extrinsic matrices $\mathcal{M}_{\rceil}$ which are oriented towards the center of the object, followed by rendering 32 views using a ray tracing engine. \method-NVS is evaluated on Google Scanned Objects (GSO)~\cite{gso}, which is a dataset of high-quality scanned household objects. We evaluate the models extensively with three metrics covering different aspects of image similarity: PSNR, SSIM, and LPIPS. More details are in~\supp.

\noindent\textbf{I2I $v.s.$ T2I Prior on NVS Task.} 
To validate the effectiveness of I2I prior on the novel view synthesis task, we compare with training from scratch, I2I prior, and T2I prior on the novel view synthesis task. 
For fair comparison, we use the SOTA method, \ie, PixArt-$\alpha$~\cite{pixart-a} model, as T2I prior, whose structure is similar to our model.
As shown in~\Cref{fig:ablate_nvs}, the model transferred from I2I prior is better than from no prior and T2I prior, which demonstrates the superiority of I2I prior over T2I prior.
\par

\noindent\textbf{Comparison with SOTA NVS models.} 
We compare \method-NVS with the existing state-of-the-art models in ~\Cref{tab:SOTA_nvs_table}.
We uniformly render photos at 16 angles around the object at two elevations of 0$^{\circ}$ and 30$^{\circ}$ for testing. The comparison results show that benefiting from the I2I prior, \method-NVS outperforms the current state-of-the-art models.

\begin{table}[htbp]
    \centering
    \scriptsize
    \caption{
        \textbf{Comparison with SOTA models on novel view synthesis task.} 
    }
    \vspace{-2mm}
    \SetTblrInner{rowsep=0.5pt}      
    \SetTblrInner{colsep=4.2pt}      %
    \begin{tblr}{
        cells={halign=c,valign=m},  %
        cell{1}{1}={r=2}{},         %
        cell{1}{2}={c=3}{},         %
        cell{1}{5}={c=3}{},         %
        hline{6}={1-7}{},      %
        hline{1,3,7}={1.0pt},        %
        vline{2,5}={1-9}{},       %
        column{1}={halign=l},
    }
        Method&Elevation 0$^{\circ}$ &&& Elevation 30$^{\circ}$&& \\
        & PSNR$\uparrow$ & SSIM$\uparrow$ & LPIPS$\downarrow$ & PSNR$\uparrow$ & SSIM$\uparrow$ & LPIPS$\downarrow$ \\
        Zero123~\cite{zero123} & 17.73 & 0.8115 & 0.1763 & 18.12 & 0.8099 & 0.1584\\
        Zero123-XL~\cite{zer123xl} & 17.68 & 0.7984 & 0.1988 & 19.20 & 0.8210 & 0.1521 \\
        SyncDreamer~\cite{syncdreamer} & - & - & - & 18.98 & 0.8284 & 0.1535 \\
        \method-NVS & \textbf{19.63} & \textbf{0.8439} & \textbf{0.1526} & \textbf{20.28} & \textbf{0.8419} & \textbf{0.1317} \\
    \end{tblr}
    \vspace{-2mm}
    \label{tab:SOTA_nvs_table}
\end{table}

\begin{figure}[htbp]
    \centering
    \includegraphics[width=\linewidth]{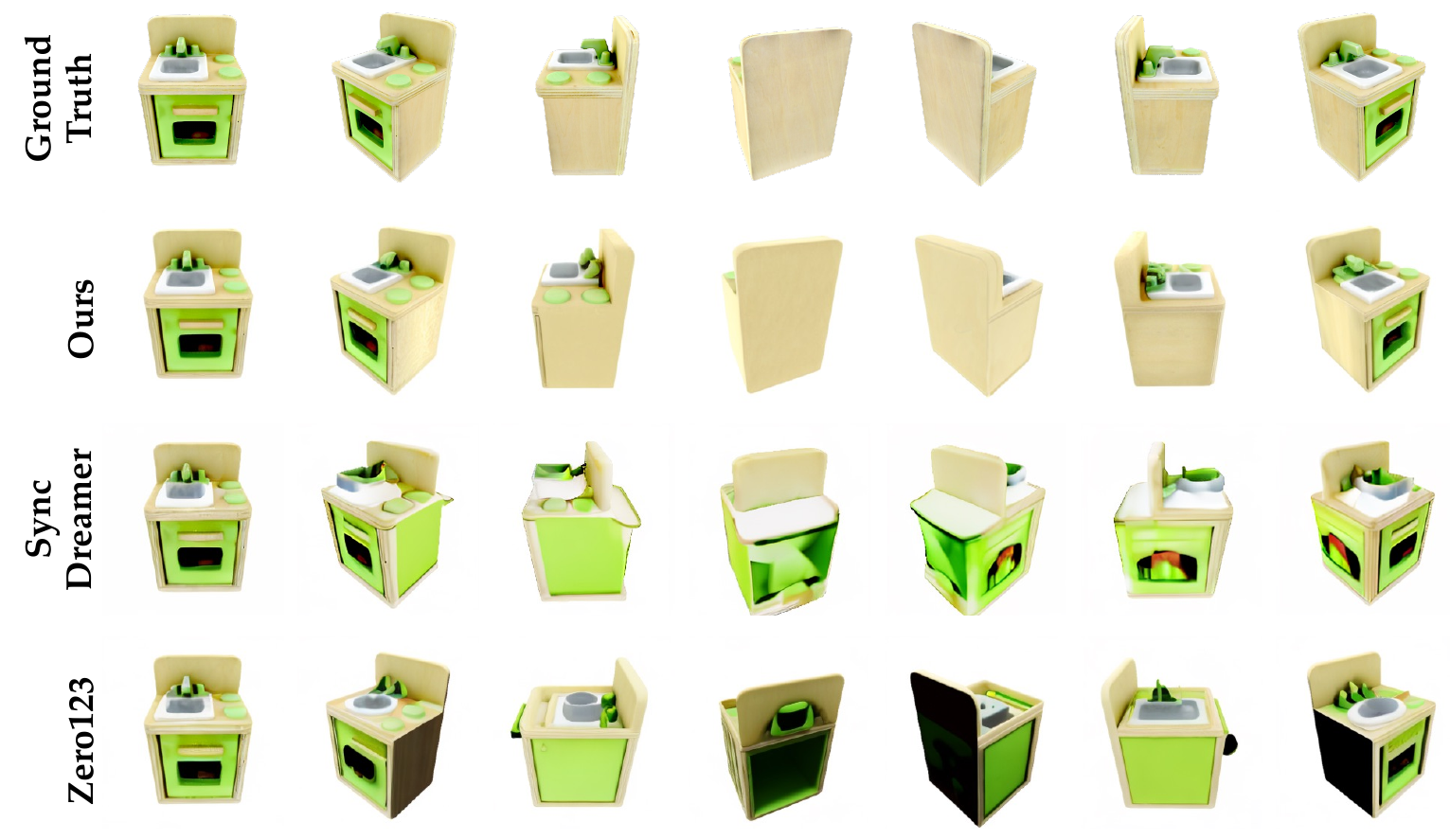}
    \vspace{-18pt}
    \caption{\textbf{Comparison of qualitative results between \method-NVS with the SOTA models.}} 
    \vspace{-10pt}
    \label{fig:qualitative_nvs}
\end{figure}

\noindent\textbf{Qualitative Results.} \Cref{fig:qualitative_nvs} illustrates a qualitative comparison of novel view generation results for a GSO test object. As can be seen, the novel view images \method-NVS generated are more notably consistent and realistic. \par

\subsubsection{Image-to-Video Generation}
Following the experimental setup in Stable Video Diffusion~\cite{svd}, we transfer \method-I2I to \method-I2V for the image-to-video~(I2V) generation task. The image-to-video task takes a still-input image as a conditioning input to generate subsequent continuous video frames.\par
\noindent\textbf{Implementation Details \& Datasets.} We use the common open-source video datasets Webvid-10M~\cite{webvid} and a subset of Openvid-1M~\cite{openvid}, which contains approximately 0.4M videos. For evaluating the performance of I2V models, we report the Fréche Video Distance (FVD)~\cite{fvd}, the Kernel Video Distance (KVD)~\cite{fvd}, and the Perceptual Input Conformity (PIC)~\cite{dynamicrafter} on UCF-101~\cite{ucf101} and MSR-VTT~\cite{msrvtt}. More details are in~\supp.

\noindent\textbf{I2I $v.s.$ T2I Prior on I2V Task.} We compare I2I and T2I prior for the image-to-video generation task through zero-shot evaluating on UCF-101~\cite{ucf101} and MSR-VTT~\cite{msrvtt} datasets. 
For the T2I prior, the experiment setting follows the ones of the NVS experiments. 
As shown in~\Cref{tab:ablate_i2v}, T2I and I2I prior exhibit a notable initialization for the I2V task. Meanwhile, the comparative results indicate that the I2I prior holds distinct advantages in this task.

\begin{table}[htbp]
    \centering
    \scriptsize
    \caption{
        \textbf{Comparison with different priors for I2V.}
    }
    \vspace{-2mm}
    \SetTblrInner{rowsep=0.5pt}      
    \SetTblrInner{colsep=5.8pt}      %
    \begin{tblr}{
        cells={halign=c,valign=m},  %
        cell{1}{1}={r=2}{},         %
        cell{1}{2}={c=3}{},         %
        cell{1}{5}={c=3}{},         %
        hline{1,3,6}={1.0pt},        %
        vline{2,5}={1-9}{},       %
    }
        Method & UCF-101~\cite{ucf101} & & & MSR-VTT~\cite{msrvtt} & & \\
               & FVD$\downarrow$ & KVD$\downarrow$ & PIC$\uparrow$ & FVD$\downarrow$ & KVD$\downarrow$ & PIC$\uparrow$ \\
        $w/o$ Prior & 2892.09 & 170.81 & 0.3795 & 1490.64 & 72.97 & 0.3999 \\
        $w$ T2I Prior & 532.77 & 52.53 & 0.7321 & 313.50 & \textbf{19.61} & 0.7321 \\
        $w$ I2I Prior & \textbf{399.59} & \textbf{39.17} & \textbf{0.7930} & \textbf{271.00} & 21.16 & \textbf{0.7869} \\ 
    \end{tblr}
    \vspace{-2mm}
    \label{tab:ablate_i2v}
\end{table}

\subsection{Ablation Study} \label{ablation study}
Finally, we conduct a comprehensive series of experiments to thoroughly analyze and evaluate the effectiveness of various generative priors and model components. 

\begin{figure}[htbp]
\centering
\subfloat[]{
 \centering
 \includegraphics[width=0.45\linewidth]{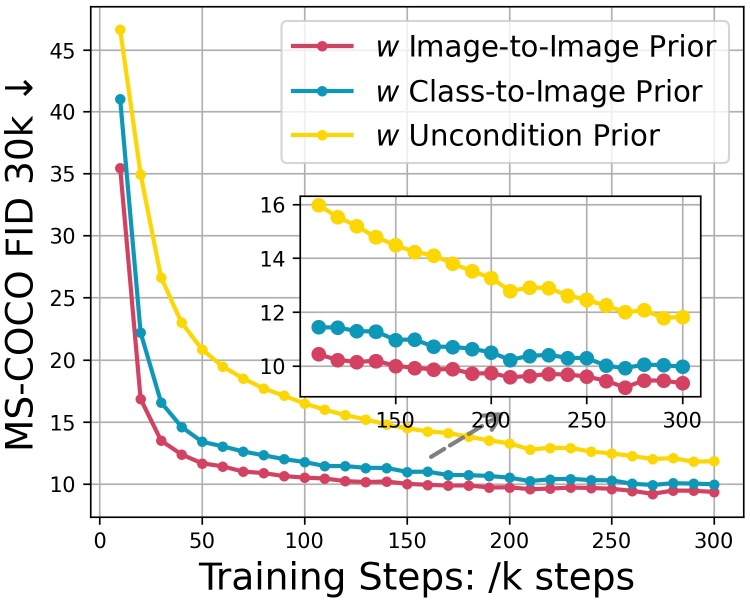}
}
\subfloat[]{
 \centering
 \includegraphics[width=0.45\linewidth]{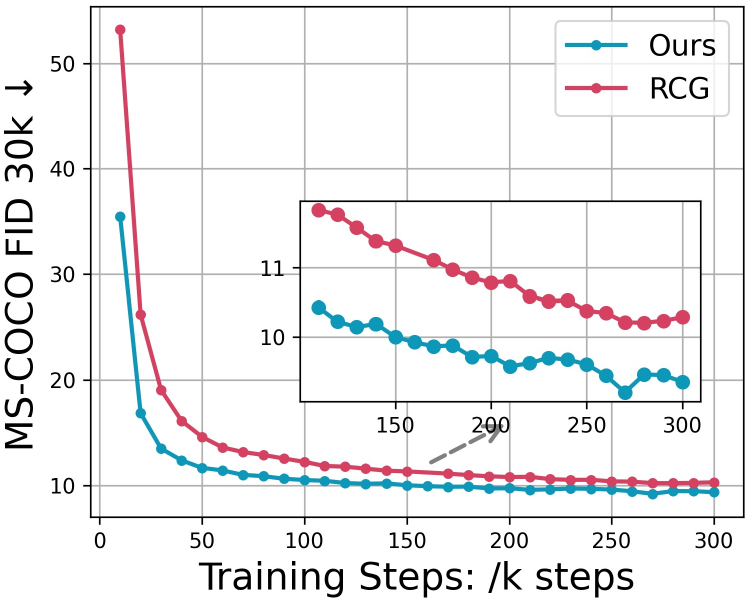}
}
\vspace{-10pt}
\caption{(a) Comparison with different priors of I2I generation on T2I task, (b) Comparison our I2I model with RCG on T2I task.}
\label{fig:ablate pretrain task}
\vspace{-10pt}
\end{figure}

\noindent\textbf{Ablating different generative priors.} 
We compare \method-I2I prior with the class-to-image prior and uncondition prior. We also conduct a comparison with I2I prior from RCG~\cite{RCG2023}, which uses adaLN-Zero block for injecting conditions.
To ensure fair experiments, all frameworks are pre-trained on ImageNet-1K~\cite{deng2009imagenet} and fine-tuned on the same image-text dataset. 
As shown in~\Cref{fig:ablate pretrain task} (a) and (b), \method-I2I provides better prior knowledge for downstream the T2I generation task.

\begin{figure}[htbp]
    \centering
    \subfloat[FID score on I2I task]{
     \centering
     \includegraphics[width=0.45\linewidth]{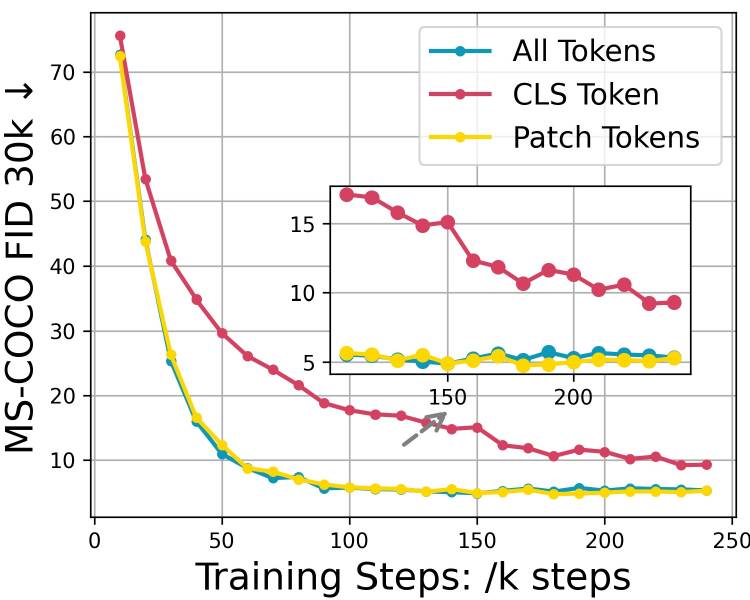}
    }
    \subfloat[FID score on T2I task]{
     \centering
     \includegraphics[width=0.45\linewidth]{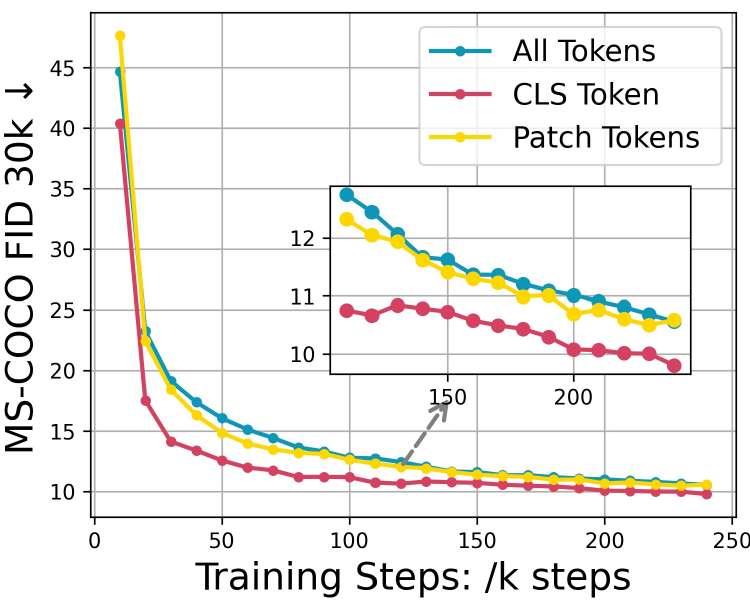}
    }
    \vspace{-10pt}
\caption{Comparison with different condition features of CLIP.}
\label{fig:ablate token type}
\vspace{-12pt}
\end{figure}
\noindent\textbf{Ablating different features for I2I generative prior.}
To explore which types of visual features (global semantic or fine-grained local information) are more suitable for the I2I framework, we set up three comparative experiments, \ie, a single global token ([\texttt{CLS}] token), local patch tokens, and all vision tokens (single global token + local patch tokens). As shown in~\Cref{fig:ablate token type} (a) and (b), sufficient fine-grained local information significantly accelerates the convergence speed during the I2I training process. However, it is interesting that fine-grained local features lead to a high dependence of the pre-trained model on the conditions, which is not conducive to the rapid transfer of the pre-trained model to downstream tasks. \label{question1} \par

\begin{figure}[htbp]
    \centering
    \subfloat[]{
     \centering
     \includegraphics[width=0.45\linewidth]{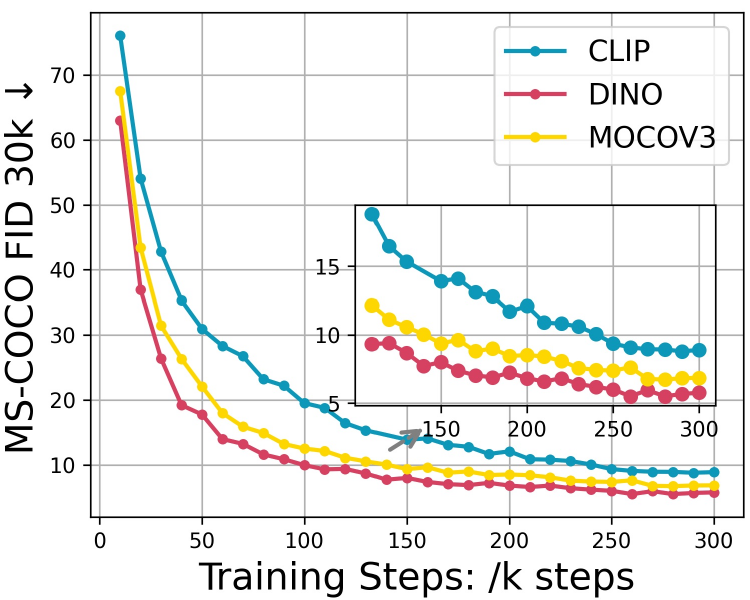}
    }
    \subfloat[]{
     \centering
     \includegraphics[width=0.45\linewidth]{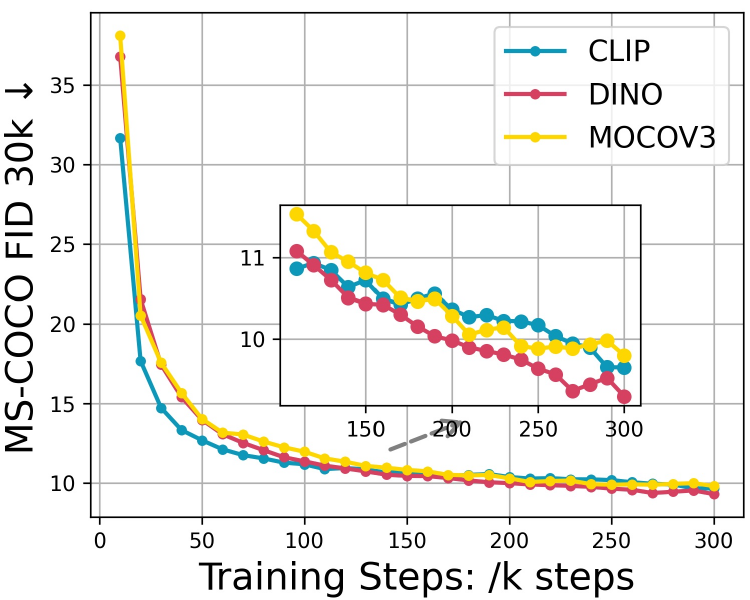}
    }
    \vspace{-11pt}
\caption{(a) Comparison with different vision encoders on I2I task, (b) Comparison with different vision encoders on downstream T2I task.}
\label{fig:ablate vision encoder}
\vspace{-15pt}
\end{figure}
\begin{figure}[htbp]
    \centering
    \subfloat[I2I with CLIP-L on T2I]{
         \centering
         \includegraphics[width=0.45\linewidth]{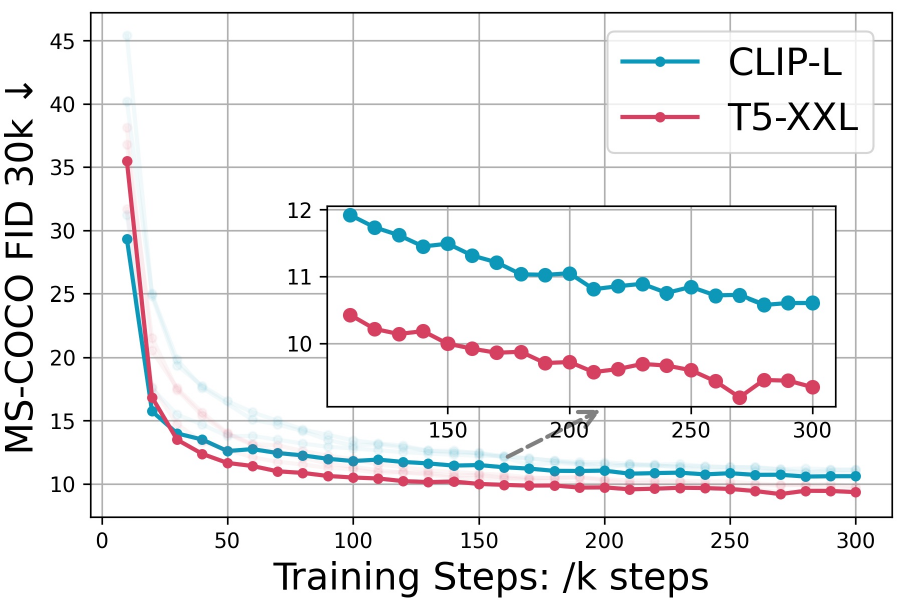}
    }
    \subfloat[I2I with DINO-B on T2I]{
         \centering
         \includegraphics[width=0.45\linewidth]{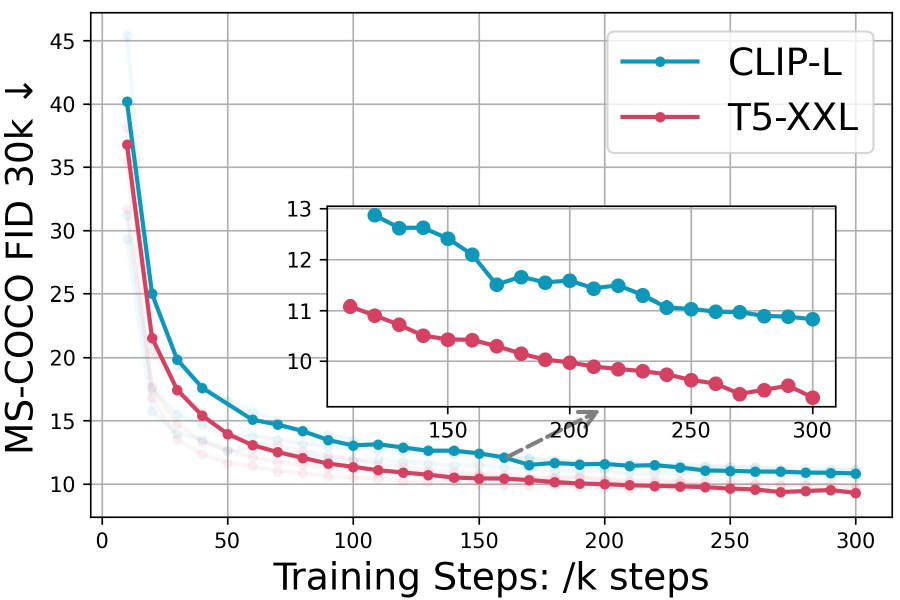}
     }
    \vspace{-10pt}
\caption{Comparison between T5 and CLIP text encoder.}
\label{fig:ablate text encoder}
\vspace{-11pt}
\end{figure}
\noindent\textbf{Ablating vision encoders for I2I generative prior.} 
We set up experiments to probe the impact of different types of vision encoders (\eg, unimodal self-supervised $v.s.$ multimodal alignment) on our I2I frameworks. 
We adopt three types of vision encoders: CLIP, DINO, and MoCoV3. 
The comparison results in~\Cref{fig:ablate vision encoder} (a) demonstrate that the self-supervised pre-trained vision encoder (\ie, DINO and MoCoV3) has a significantly faster convergence speed in the image-to-image task. 
We then validate the transfer performance on the downstream task that relies most on text-image alignment (text-to-image generation). The comparison results in~\Cref{fig:ablate vision encoder} (b) show that the image-text alignment models have significant advantages during the initialization phase of transfer learning. However, this advantage quickly diminishes during the fine-tuning process. Thus, I2I generation framework has a robustness for different vision encoder types.\label{question2} \par

\begin{figure}[htbp]
    \centering
    \subfloat[]{
         \centering
         \includegraphics[width=0.45\linewidth]{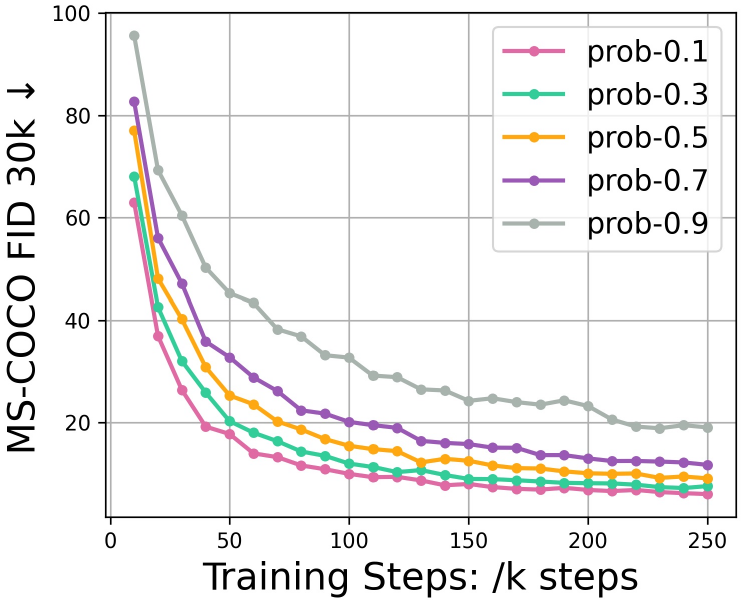}
    }
    \subfloat[]{
         \centering
         \includegraphics[width=0.45\linewidth]{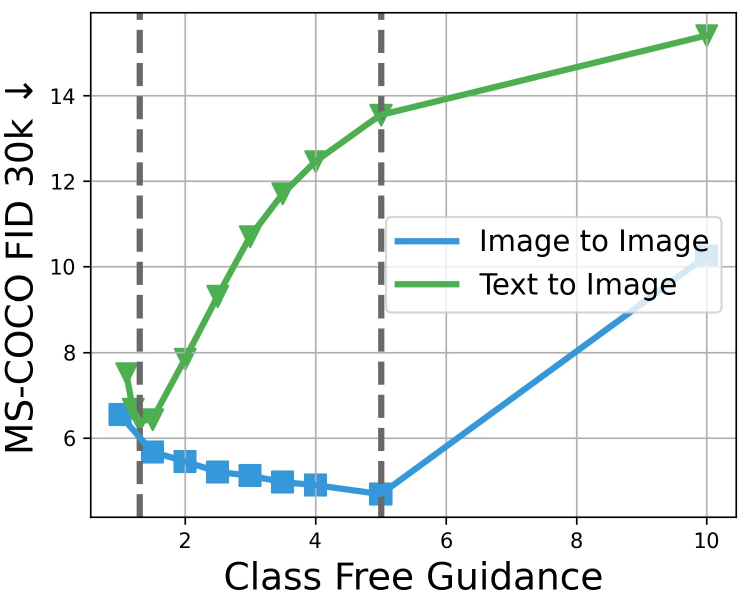}
    }
    \vspace{-10pt}
\caption{(a) Different condition dropout probability during training, (b) Different class-free guidance factor during inference.}
\label{fig:ablate componment}
\vspace{-11pt}
\end{figure}

\noindent\textbf{Ablating different text Encoders for T2I task.} We compare the performance of T5 and CLIP text encoder on different I2I generative priors, as shown in~\Cref{fig:ablate text encoder} (a) and (b). For the I2I generative priors with unimodal vision encoders (\eg, DINO), T5 outperformed the CLIP text encoder in both convergence speed and performance in terms of  FID. For priors with multimodal vision encoders (\eg, CLIP-L), the corresponding CLIP text encoder shows greater advantages in the early transferring stage. However, T5 finally achieves better performance.

\noindent\textbf{Ablating class-free guidance for I2I model.} We ablate condition dropout probability during the training process and the class-free guidance factor in the inference phrase in~\Cref{fig:ablate componment} (a) and (b). We investigate that the condition dropout probability during training exhibits similar properties in image-to-image tasks as in text-to-image tasks. However, for the class-free guidance factor, we observe that image generation tasks have a broader applicability and a significant robustness advantage. \par

%% file: sec/5_conclusion.tex
\section{Conclusion}\label{sec:conclusion}

In this paper, we probe the image-to-image (I2I) generation, where models can learn from in-the-wild images in a self-supervised manner. 
First, we develop a purely vision-based training framework, \method, and validate the feasibility and scalability of learning I2I models. 
Our findings reveal that, as an upstream task of text-to-image (T2I) generation, our I2I model provides a more fundamental visual prior, achieving comparable or superior performance to current T2I models.
It is worth mentioning that we only utilize 1/10 of the text-image pairs for fine-tuning of T2I models.
Furthermore, we demonstrate the advantages of I2I priors over T2I priors in text-irrelevant visual generative tasks, such as image-to-3D and image-to-video generation.
We believe this approach has the potential to liberate image generation from the constraints of image-text pairs, allowing it to learn a more foundational pure-visual prior.

%% file: sec/Acknowledgement.tex
\section*{Acknowledgment}\label{sec:acknowledgment}
This work is supported by National Nature Science Foundation of China (grant No.61871106), Key R\&D projects of Liaoning Province, China (grant No.2024JH2/102500015). This work was supported by Ant Group Research Intern Program.

%% file: sec/6_ref.tex
\small
\bibliographystyle{ieeenat_fullname}
\bibliography{main}

\clearpage
\newpage

%% file: sec/7_appendix.tex
\renewcommand\thesection{\Alph{section}}
\renewcommand\thefigure{S\arabic{figure}}
\renewcommand\thetable{S\arabic{table}}
\renewcommand\theequation{S\arabic{equation}}

\setcounter{figure}{0}
\setcounter{table}{0}
\setcounter{equation}{0}
\setcounter{page}{1}

\onecolumn
{\centering
\Large
\textbf{\thetitle}\\
\vspace{0.5em}Supplementary Material \\
\vspace{1.0em}}

\newcommand\DoToC{%
    \hypersetup{linkcolor=blue}
    \startcontents
    \printcontents{}{1}{\hrulefill\vskip0pt}
    \vskip0pt \noindent\hrulefill
    \hypersetup{linkcolor=red}
    }

\noindent\DoToC 

\begin{wrapfigure}{r}{0.3\textwidth}
\vspace{-10pt}
\centering
\includegraphics[width=\linewidth]{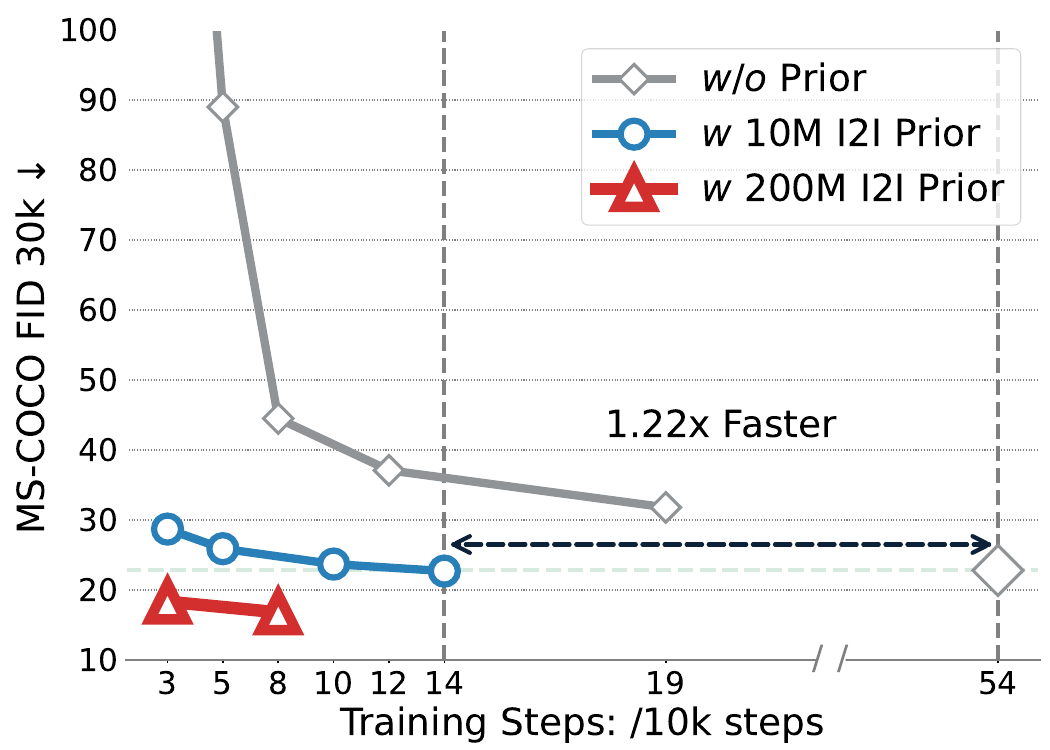}
\vspace{-10pt}
\caption{Supplementary FID results for motivation experiments.}
\label{fig:FID Moti}
\vspace{-30pt}
\end{wrapfigure}

\section{Supplementary Experiments}
\subsection{Motivation experiments FID results}
As shown in \Cref{fig:FID Moti}, we provide the FID metric results of the motivation experiment (Figure \red{1b}). Although the quality measurement standard of image-text pairs is customized based on clip score, our \method training framework also shows significant advantages in FID metric.

\subsection{Ablate the number of condition tokens for I2I framework}
\begin{wrapfigure}{r}{0.6\textwidth}
\vspace{-10pt}
\centering
\subfloat[]{
     \centering
     \includegraphics[width=0.48\linewidth]{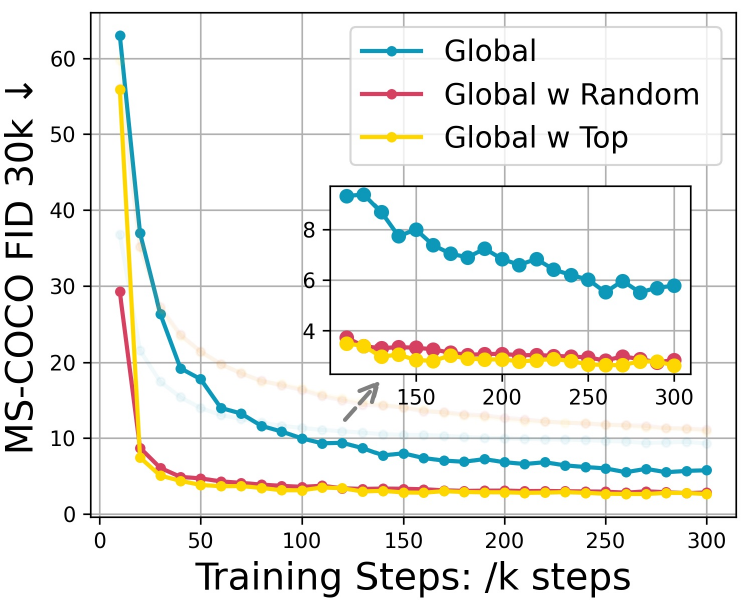}
}
\subfloat[]{
     \centering
     \includegraphics[width=0.48\linewidth]{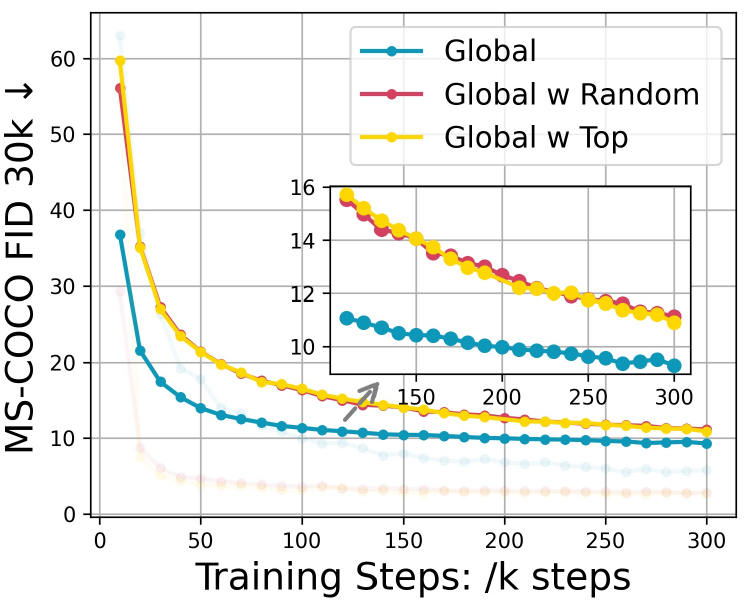}
}
\vspace{-10pt}
\caption{(a) Image-to-Image Generation, (b) Text-to-Image Generation.}
\label{fig:ablate num}
\vspace{-20pt}
\end{wrapfigure}

We explore this issue by setting the same number of condition tokens during the I2I framework as in the downstream text-to-image tasks. For the text-to-image task, we used T5~\cite{T5}. For the I2I training, we set the number of condition tokens to be the same as in the downstream task, which is $120$. We conduct experiments using DINO-B, leveraging the excellent ability of the DINO [\texttt{CLS}] token to focus on foreground tokens. We select the top 119 tokens with a high correlation with the class token. Additionally, we design a set of comparative experiments, randomly selecting $119$ tokens from patch tokens to introduce noise perturbation. The experimental results in~\Cref{fig:ablate num} (a) and (b) indicate that more fine-grained local information accelerates the training of I2I generation. However, the standalone global class token still exhibits superior transfer performance for downstream tasks.

\subsection{Ablating the Scale of Vision Encoder} 
\begin{wrapfigure}{r}{0.6\textwidth}
\vspace{-10pt}
\centering
\subfloat[]{
     \centering
     \includegraphics[width=0.48\linewidth]{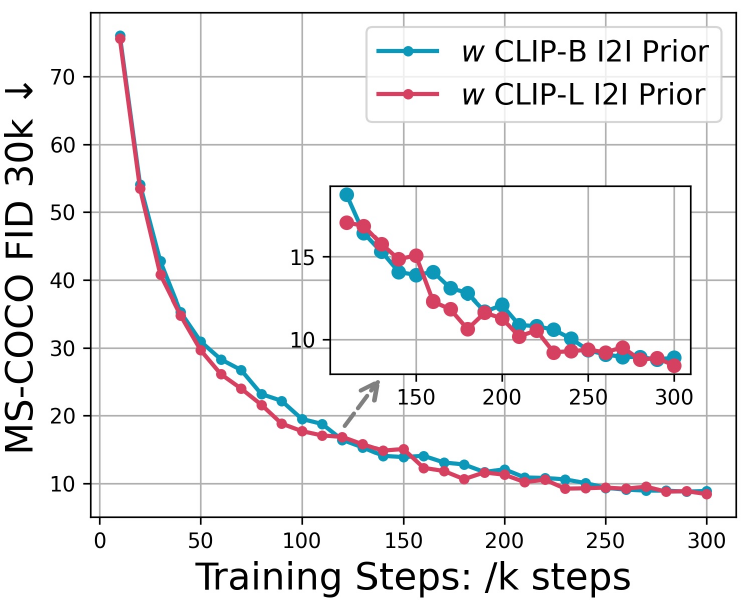}
 }
\subfloat[]{
     \centering
     \includegraphics[width=0.48\linewidth]{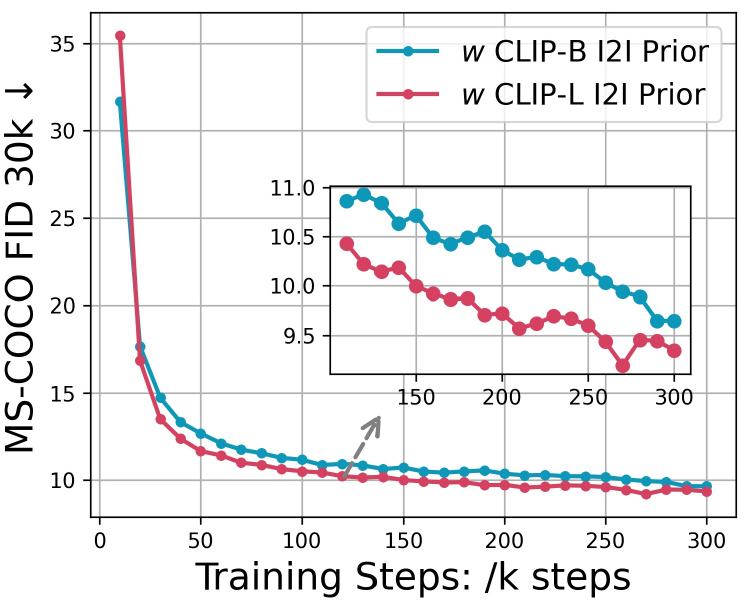}
}
\vspace{-10pt}
\caption{(a) Image-to-Image Generation, (b) Text-to-Image Generation.}
\label{fig:ablate scale}
\vspace{-30pt}
\end{wrapfigure}

The comparison results in \Cref{fig:ablate scale} (a) and (b) demonstrate that larger and better vision encoders can provide a higher performance ceiling for downstream tasks. Therefore, for a better vision encoder, \method-I2I framework has greater potential.

\section{Dataset}
This section supplements a detailed introduction of training data and implementation details for the models.
\subsection{Image-to-Image Generation}
The training data of \method-I2I can be expanded indefinitely. In this paper, we curate and construct a pure image dataset totaling 190 million from existing open-source data. This includes 120 million images filtered from LAION-5B~\cite{laion} with a resolution greater than 512 and an aesthetic score greater than 5.0, as well as 55 million images selected from COYO-700M~\cite{coyo-700m} with the same resolution and aesthetic score criteria. Additionally, we include 10 million high-quality segmented scene data from the SAM~\cite{sam} dataset, 4 million high aesthetic score images from JourneyDB~\cite{journeydb}, and 1 million classic natural scene images from the ImageNet-1K~\cite{deng2009imagenet} dataset.\par

\subsection{Text-to-Image Generation}
We construct a dataset of 30 million text-image pairs, all of which are sourced from easily accessible open datasets. This includes 10 million and 5 million images selected from LAION-5B~\cite{laion} and COYO-700M~\cite{coyo-700m} respectively, based on a standard resolution greater than 512 and an aesthetic score higher than 5.5. Additionally, the dataset contains 10 million from SAM~\cite{sam}, 4 million from JourneyDB~\cite{journeydb}, and 1 million from Imagenet-1K~\cite{deng2009imagenet}. For the text caption, we use the state-of-the-art multi-modal large language model (\ie, InternVL~\cite{internvl}) to generate detailed long captions. Following DALLE-3~\cite{dalle3}, we incorporate raw captions into the training process.

\subsection{Novel View Synthesis}
We finetune \method-I2I for novel view synthesis task using a subset (750k) of the released Objaverse~\cite{objaverse} dataset, a large-scale open-source collection comprising over 800K 3D models created by more than 100K artists. We randomly sample 32 camera extrinsic matrices $\mathcal{M}_{\rceil}$ which are oriented towards the center of the object, followed by rendering 32 views using a ray tracing engine. 

\subsection{Image-to-Video Generation}
For the image-to-video generation task, \method-I2V is initialized from \method-I2I and trained on WebVid10M~\cite{webvid} dataset by sampling 16 frames with 3 frames interval.

\section{Model and Implementation Details}
\subsection{\method-I2I Model}
\noindent\textbf{Implementation and Training Details.} We train the \method-I2I (as shown in \Cref{method} (a)) on 64 A100 GPUs with the total batch size of 16384. For saving memory, we use the mixed $fp16$ format with gradient checkpointing. The AdamW optimizer is utilized with a weight decay of 0.03 and a constant $1.6\times10^{-4}$ learning rate. In the initial phase of training, we set a warm-up of 1000 steps for stable training.\par

\subsection{\method-T2I Model}
\noindent\textbf{Implementation and Training Details.} \method-T2I (as shown in \Cref{method} (b)) uses the T5~\cite{T5} large language model (specifically 4.3B Flan-T5-XXL) as the text encoder for conditional feature extraction, with the text condition length set to 120. Inspired by SDXL~\cite{sdxl} and Pixart-$\alpha$~\cite{pixart-a}, \method-T2I adopts the progressive resolution training strategy and provide four resolution versions of the text-to-image model, which are 256$\times$256, 512$\times$512, 1024$\times$1024, and an arbitrary aspect ratio version at 1024 resolution. \method-T2I is trained on 64 A100 GPUs. Following the training setting of \method-I2I, we use the mixed $fp16$ format with gradient checkpointing with the AdamW optimizer and the warm-up setting for stable training. Details of the training information are shown in \Cref{Tab:lumos-t2i training details}.

\begin{table*}[htbp]
\centering
\caption{\textbf{Detailed training information about every \method-T2I training stage.}}\label{Tab:lumos-t2i training details}
\resizebox{0.8\textwidth}{!}{
\begin{tabular}{ccccccc}
\toprule
Stage & Image Resolution & Training Steps(K) & Batch Size & Learning Rate & weight decay & Warm Up Steps(K) \\ \midrule
1     & 256$\times$256          & 65                & 256$\times$64     & 1.6$\times$10$^{-4}$      & 0.03         & 1                \\
2     & 512$\times$512          & 60                & 64$\times$64      & 8$\times$10$^{-5}$       & 0.03         & 1                \\
3     & 1024$\times$1024        & 20                & 16$\times$64      & 4$\times$10$^{-5}$        & 0.03         & 1                \\
4     & Multi-scale 1024 & 20                & 16$\times$64      & 4$\times$10$^{-5}$        & 0.03         & 1                \\ \bottomrule
\end{tabular}}
\end{table*}

\subsection{\method-NVS Model}
\noindent\textbf{Training Objective.}
Inspired by the definition of the task in Zero-1-to-3~\cite{zero123}, we tune \method-I2I to the novel view synthesis task. 
As shown in \Cref{method} (c), the task is to synthesize an image of an object from a new camera viewpoint.
The training data consists of image-viewpoint pairs, where each pair includes a single image $x \in \mathbb{R}^{h \times w \times 3}$ and its corresponding condition $c = (R,T)$.
In detail, the relative camera rotation $R \in \mathbb{R}^{3 \times 3}$ and translation $T \in \mathbb{R}^{3}$ determine the desired viewpoint. 
\method-NVS is trained via
\begin{equation}
L_{\theta_{\text{NVS}}}:=\mathbb{E}_{\mathcal{E}(\hat{x}_{R, T}),\mathcal{E}(x), x, \epsilon \sim \mathcal{N}(0,1), t}\left[\left\|\epsilon-\epsilon_\theta\left(\left<z_t,\mathcal{E}(x)\right>, t, \tau^{img}(x),R,T\right)\right\|_2^2\right],
\end{equation}
where $\left<\cdot, \cdot\right>$ represents the concatenate operation and $\hat{x}_{R, T}$ denotes the synthesized image. Meanwhile, based on the generated novel view image list, we can directly use the off-the-shelf sparse views 3D Reconstructor (\eg, LGM~\cite{lgm} and GRM~\cite{grm}) to handle the Single View 3D Reconstruction task. In this paper, we mainly use the open-source LGM to reconstruct the new perspective of the sparse view generated by our model into a 3D Gaussian~\cite{gaussian}.

\noindent\textbf{Implementation and Training Details.}
We train \method-NVS on 64 A100 GPUs with a total batch size of 16384. The mixed $fp16$ format with gradient checkpointing is utilized for saving memory. The AdamW optimizer is utilized with a weight decay of 0.03 and a constant $1.6\times10^{-4}$ learning rate. We set a warm-up of 1000 steps for stable training in the initial phase of training.\par

\subsection{\method-I2V Model}
\noindent\textbf{Training Objective.}
We finetune our \method-I2I for the image-to-video generation task, where the video model receives a still input image as the condition.
Following stable video diffusion~\cite{svd}, we use a 2D VAE encoder $\mathcal{E}$ to compress each frame of an n-frame video $v = [f^{0},\dots,f^{n}]$ into latent representation $z^{[1\dots n]} = [z^{0},\dots,z^{n}],$ where $z^{i} = \mathcal{E}(f^{i})$.
Besides the diffusion transformer initialized from \method-I2I, we attach the temporal module to the \method-I2V as shown in \cref{method} (d).
Transformers can be easily extended to support image-to-image and video-to-video tasks due to the macro modeling ability.
Different from SVD, which concatenates the condition frame to the latent noise of all generation video frames, we leverage mask strategy~\cite{zheng1open}) to support image conditioning. 
Meanwhile, we maintain the original image condition control method of \method-I2I for the image-to-video task. 
The training objective is as follows:
\begin{equation}
L_{\theta_{\text{I2V}}}:=\mathbb{E}_{[\mathcal{E}(f^{0}),\mathcal{E}(f^{1}),\dots,\mathcal{E}(f^{n})], f_{0}, \epsilon \sim \mathcal{N}(0,1), t}\left[\left\|\epsilon-\epsilon_\theta\left([z^0_{0},z^1_{t},\dots, z^n_{t}], t, \tau^{img}(f_{0})\right)\right\|_2^2\right].
\end{equation}
where $c_{\text{in}}$ includes the latent representation of the first frame $z^{0}$ and the extracted semantic information of the first frame $\tau^{img}$($f^{0}$).
During the inference phase, we unmask the conditional frame. Moreover, the unmasked frame is assigned timestep 0, while others remain the same $t$.

\noindent\textbf{Implementation and Training Details.}
We train \method-I2V on 64 A100 GPUs with a total batch size of 4096. We use $bf16$ format with gradient checkpointing and ZeRO stage-2 optimizer. The HybridAdam optimizer is set with $2\times10^{-5}$ learning rate. \par

\begin{figure*}[htbp]
    \centering
    \includegraphics[width=0.9\textwidth]{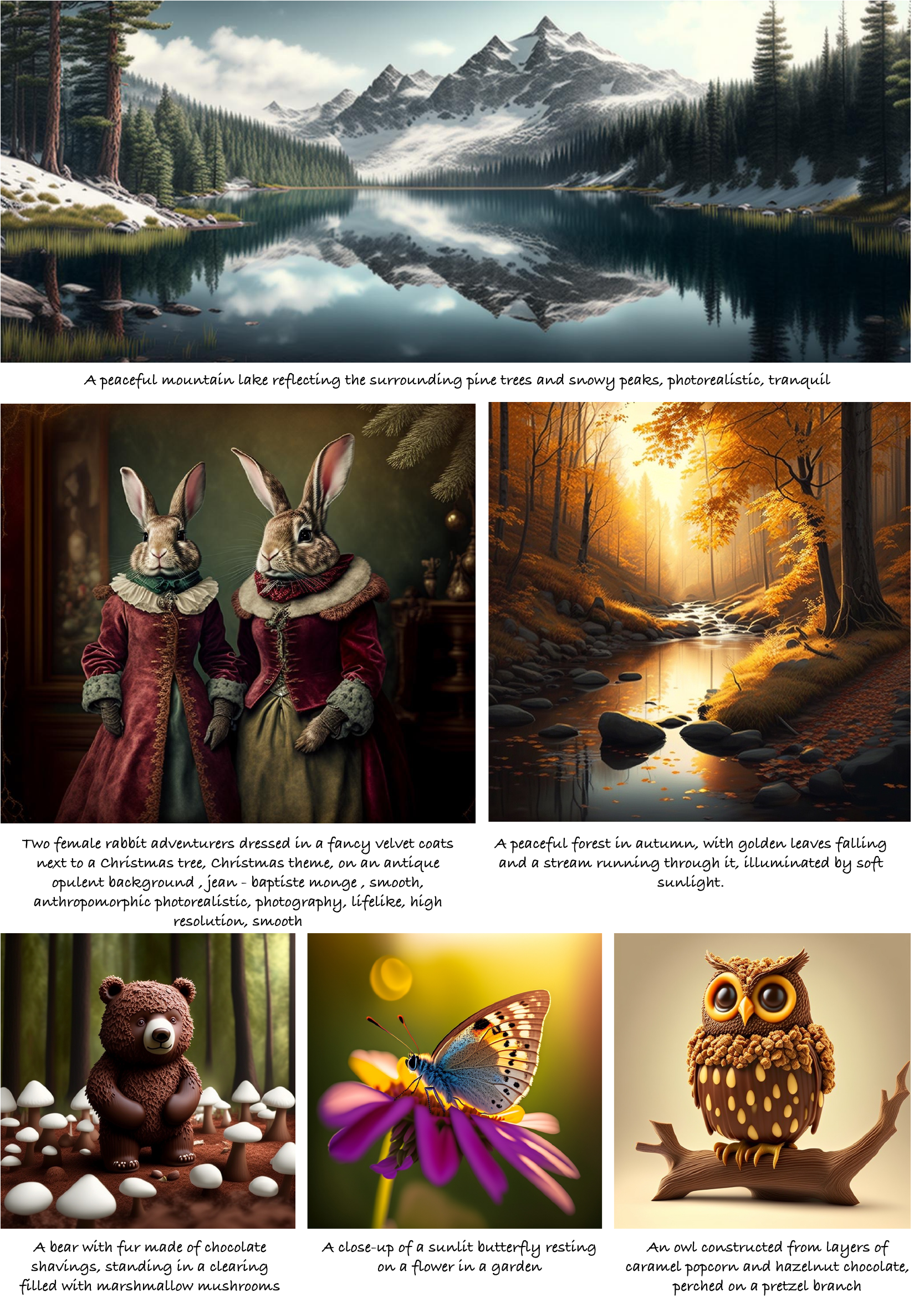}
    \vspace{-10pt}
    \caption{The samples generated by \method-T2I exhibit remarkable quality, characterized by exceptional fidelity and precise alignment with the provided textual descriptions.}
    \label{fig:qualtative1}
\end{figure*}

\begin{figure*}[htbp]
    \centering
    \includegraphics[width=0.9\textwidth]{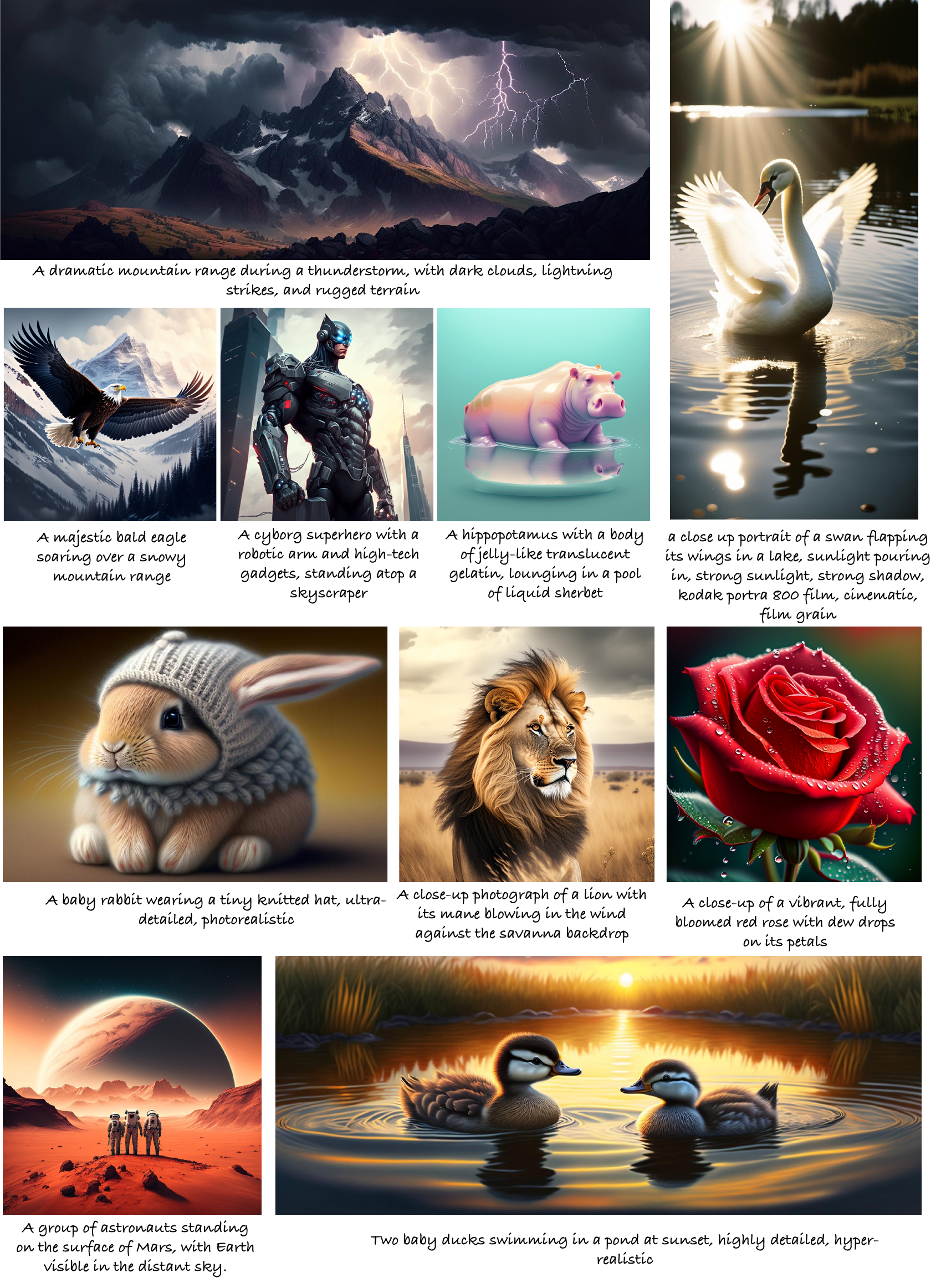}
    \vspace{-10pt}
    \caption{The samples generated by \method-T2I exhibit remarkable quality, characterized by exceptional fidelity and precise alignment with the provided textual descriptions.}
    \label{fig:qualtative2}
\end{figure*}

\begin{figure*}
    \centering
    \includegraphics[width=\linewidth]{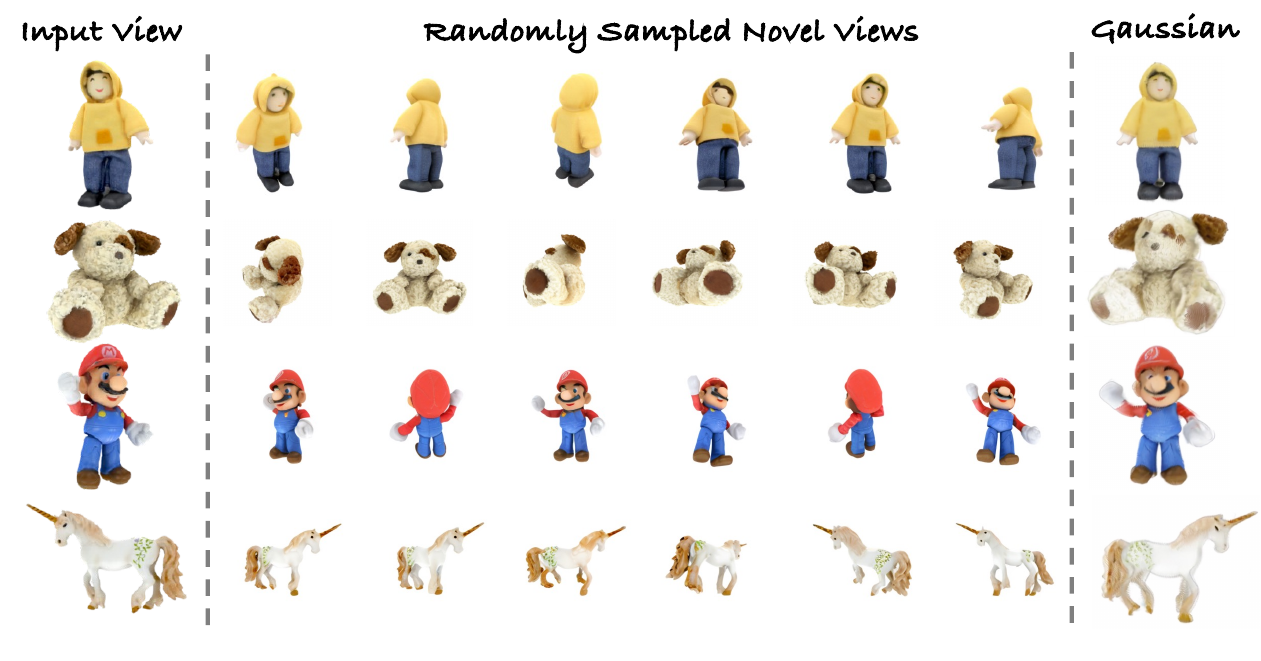}
    \caption{\textbf{Qualitative results of \method-NVS}, where the leftmost one is the input view, the middle ones are the randomly sampled generated views, and the rightmost one is the reconstructed Gaussian.}
    \label{fig:qualitative_nvs_appendix}
\end{figure*}

\begin{figure*}
    \centering
    \includegraphics[width=\linewidth]{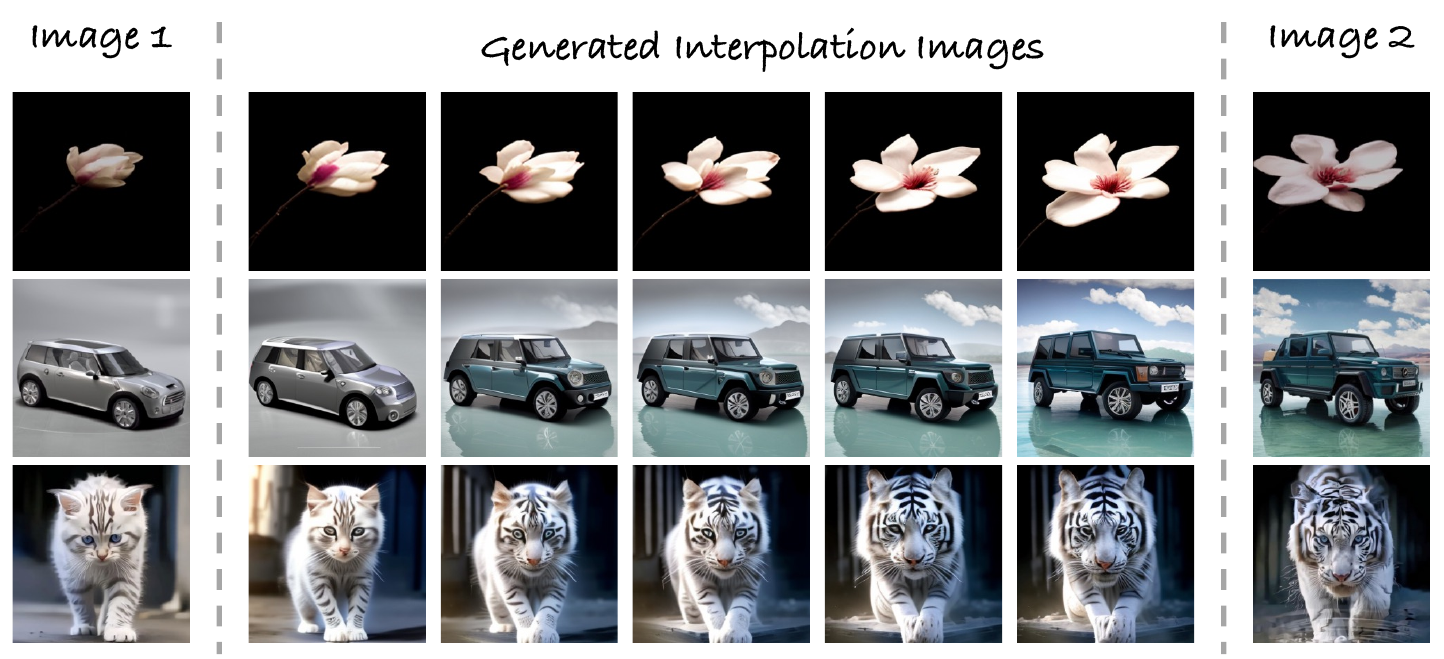}
    \caption{\textbf{Qualitative results of \method-I2I Interpolation}, where the leftmost and the rightmost ones are the input images and the middle ones are generated interpolation images.}
    \label{fig:interpolation}
\end{figure*}

\begin{figure*}
    \centering
    \includegraphics[width=\linewidth]{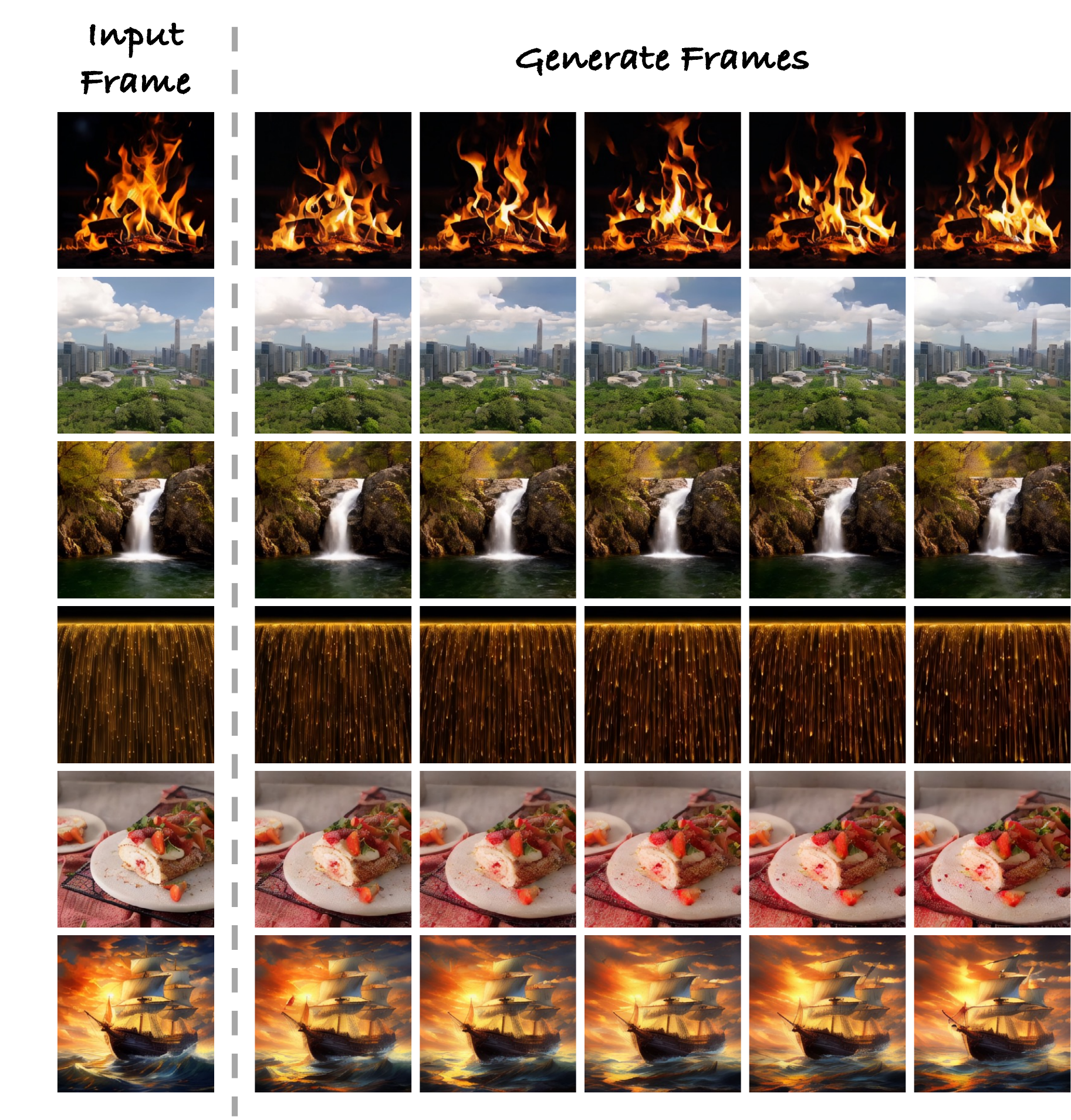}
    \caption{\textbf{Qualitative results of \method-I2V}, where the leftmost one is the input frame, right ones are generated frames.}
    \label{fig:qualitative_i2v}
\end{figure*}

\section{Qualitative Results}
This section provides more qualitative results of text-to-image generation, novel view synthesis, and image-to-video generation tasks. Moreover, we exhibit the generative capabilities of our \method-I2I for image interpolation.

\subsection{Text-to-Image Generation}
As shown in \Cref{fig:qualtative1} and \Cref{fig:qualtative2}, we provide more generated images and their corresponding text prompts. \method-T2I can generate high-quality aesthetic photos while maintaining image and text alignment.

\subsection{Novel View Synthesis}
We provide more examples and novel views generated by \method-NVS in \Cref{fig:qualitative_nvs_appendix}.
\subsection{Image-to-Video Generation}
More videos generated by \method-I2V from the input frame are provided in \Cref{fig:qualitative_i2v}.

\clearpage
\newpage
\subsection{Image Interpolation}
We provide an application example of \method-I2I in image interpolation, as shown in \Cref{fig:interpolation}. Since \method-I2I can generate images that highly retain the original image information, it has a good effect on the image interpolation task.

\section{Prompts in Figure \red{1a}}
We provide the text prompts adopted to generate images in Figure \red{1a}. The prompts are arranged from top to bottom, left to right. 
\begin{itemize}
    \item \textit{``golden sunset shines on the top of snow-capped mountains, with small villages at its foot and surrounding buildings.''}
    \item \textit{``A rustic bedroom showcasing a round bed, earth-toned decor, and a cluttered, yet charming ambiance.''}
    \item \textit{``Documentary-style photography of a bustling marketplace in Marrakech, with spices and textiles.''}
    \item \textit{``group characters from fantasy myth in the style of ori and the blind forest, riot games, ghibli, ori environment.''}
    \item \textit{``Post-Apocalyptic Wanderer, character design, style by kim jung gi, zabrocki, karlkka, jayison devadas, 8k.''}
    \item \textit{``The picture shows a cute little tiger, wearing a blue hoodie and hat, sitting on a small cardboard boat on calm water.''}
    \item \textit{``A dragon made of molten chocolate, with scales that glisten like gold leaf and eyes of crystalline sugar.''}
    \item \textit{``This professional photo from National Geography shows the subtleties in a erased face of god in the shape of the subtle cloud but we can clearly see the face of almighty god with this stormy atmosphere that is brewing in this Nevada desert, volumetric lighting, high contrast.''}
\end{itemize}